\documentclass[runningheads]{llncs}

\usepackage[mobile]{eccv}

\usepackage{eccvabbrv}

\usepackage{graphicx}
\usepackage{booktabs}
\usepackage{xcolor}  
\usepackage[accsupp]{axessibility}  % Improves PDF readability for those with disabilities.
\usepackage{graphicx}
\usepackage{amsmath,amssymb} % define this before the line numbering.
\usepackage{url}            % simple URL typesetting
\usepackage{algorithm}
\usepackage{algorithmic}
\usepackage{color}
\usepackage{soul, xcolor}
\usepackage[width=122mm,left=12mm,paperwidth=146mm,height=193mm,top=12mm,paperheight=217mm]{geometry}
\usepackage{xspace}
\usepackage{adjustbox}
\usepackage{booktabs}
\usepackage{multirow}
\usepackage{makecell}
\usepackage{xcolor}
\usepackage{subcaption}
\usepackage{enumitem}
\usepackage{pifont}
\usepackage{wrapfig}
\usepackage[table]{xcolor}

\newcommand{\sqbullet}{\ding{113}} % 可把 113 换成 110/111/112/114 试一下

\definecolor{rred}{RGB}{192, 0, 0}

\definecolor{my_green}{RGB}{51,102,0}
\definecolor{my_red}{RGB}{204, 0, 0}
\renewcommand{\checkmark}{\textcolor{my_green}{\ding{51}}} % ✔
\newcommand{\crossmark}{\textcolor{my_red}{\ding{55}}} % ✘

\definecolor{myIDBcolor}{RGB}{245,248,255}

\usepackage[pagebackref,breaklinks,colorlinks,citecolor=eccvblue]{hyperref}
\usepackage{orcidlink}

\begin{document}

% ---------------------------------------------------------------
% TODO REVIEW: Replace with your title
\title{LongVQUBench: Benchmarking Long-Term Video Quality Understanding of\\Vision-Language Models} 

% TODO REVIEW: If the paper title is too long for the running head, you can set
% an abbreviated paper title here. If not, comment out.
\titlerunning{LongVQUBench}

% TODO FINAL: Replace with your author list. 
% Include the authors' OCRID for the camera-ready version, if at all possible.
% \author{First Author\inst{1}\orcidlink{0000-1111-2222-3333} \and
% Second Author\inst{2,3}\orcidlink{1111-2222-3333-4444} \and
% Third Author\inst{3}\orcidlink{2222--3333-4444-5555}}

% % TODO FINAL: Replace with an abbreviated list of authors.
% \authorrunning{F.~Author et al.}
% % First names are abbreviated in the running head.
% % If there are more than two authors, 'et al.' is used.

% % TODO FINAL: Replace with your institution list.
% \institute{Princeton University, Princeton NJ 08544, USA \and
% Springer Heidelberg, Tiergartenstr.~17, 69121 Heidelberg, Germany
% \email{lncs@springer.com}\\
% \url{http://www.springer.com/gp/computer-science/lncs} \and
% ABC Institute, Rupert-Karls-University Heidelberg, Heidelberg, Germany\\
% \email{\{abc,lncs\}@uni-heidelberg.de}}

\author{
Arpita Nema\orcidlink{0009-0008-3158-7904} \and
Hanwei Zhu\orcidlink{0000-0002-0894-0561} \and
Xi Zhang\orcidlink{0000-0002-1993-6031} \and
Weisi Lin\orcidlink{0000-0001-9866-1947}
}

\authorrunning{A. Nema et al.}

\institute{
Nanyang Technological University, Singapore \\
\email{arpita004@e.ntu.edu.sg, \{hanwei.zhu, xi.zhang, WSLin\}@ntu.edu.sg}\\
Project page: \url{https://longvqubench.github.io}.
}

\maketitle

\begin{abstract}
    The evaluation of long-term video quality understanding remains an open challenge for large vision-language models (LVLMs). Existing video quality benchmarks predominantly focus on short clips and isolated distortions, overlooking the temporal continuity, cumulative degradation, and reasoning complexity inherent in long-duration content. To address these limitations, we present \textbf{LongVQUBench}, a comprehensive benchmark for long-term video quality understanding. LongVQUBench contains over $1{,}200$ diverse videos spanning movies, documentaries, surveillance footage, egocentric recordings, and animated content, accompanied by $1{,}500$ multiple-choice and open-ended questions for validation and testing. To assess perceptual reasoning across different temporal scopes, we introduce three progressively complex evaluation levels: (i) local event quality understanding (LQU) for analyzing localized distortions; (ii) cross-event quality reasoning (CQR) for integrating multiple degraded events; and (iii) global quality understanding (GQU) for holistic perceptual evaluation over extended durations. Furthermore, a needle distortion question-answering (NDQA) paradigm is embedded across all three levels, where spatial or temporal artifacts are sparsely inserted to probe fine-grained detection and reasoning capabilities. Extensive experiments on $14$ state-of-the-art LVLMs reveal significant performance degradation with increasing video length and reasoning depth, highlighting their limited capacity for long-range temporal integration and perceptual attribution. We envision LongVQUBench as a foundational step toward the systematic, hierarchical, and explainable evaluation of LVLMs' long-term video quality understanding.

\end{abstract}

\section{Introduction}

The rapid progress of large vision-language models (LVLMs) has revolutionized multimodal understanding by integrating visual perception with linguistic reasoning. Recent models such as GPT-5~\cite{openai2025gpt5}, LLaVA-OneVision-1.5~\cite{an2025llava_onevision_1_5}, and Qwen2.5-VL~\cite{xie2025qwen2_5vl} exhibit strong capabilities in fine-grained recognition, context-aware image description~\cite{zhu2023minigpt}, and short- and long-term video question answering~\cite{zhang2025qbench,fang2024mmb, wu2024longvideobench}, signaling a transition from perception-driven analysis to semantic comprehension. Nevertheless, long-term video quality understanding (LVQU) remains underexplored. Unlike conventional video understanding tasks centered on events and semantics, LVQU requires reasoning over perceptual fidelity, temporal coherence, and cumulative degradation across extended time spans. This is an essential ability for assessing the stability and human-perceived quality of real-world long-form videos, yet still beyond the reach of current LVLMs.

Video quality assessment (VQA) provides a complementary foundation for perceptual modeling by quantifying human judgments of distortions such as compression artifacts~\cite{wang2004video,seshadrinathan2011motion,xia2009perceivable,lin2020pea265,zhang2023lvqac,zhang2021attention},
transmission errors~\cite{pinson2004new,kanumuri2006modeling,seshadrinathan2010study,yim2011evaluation}, and temporal flicker~\cite{zheng2024video,choi2016flicker,choi2018video}. Classical datasets, including LIVE-VQA~\cite{seshadrinathan2010study}, KoNViD-1k~\cite{hosu2017konvid1k}, and YouTube-UGC~\cite{wang2019youtube}, consist mainly of short clips with controlled distortions, which are valuable for low-level modeling but insufficient for capturing the temporal continuity and contextual dynamics of long videos. In contrast, LVLM-oriented benchmarks have emphasized semantic video understanding, such as action recognition and multimodal reasoning~\cite{fang2024mmb, videomme2024}. Although LongVideoBench~\cite{wu2024longvideobench} extends evaluation to long-form semantics, it remains focused on event comprehension rather than perceptual quality, while Q-Bench-Video~\cite{zhang2025qbench} explores video-quality reasoning but is restricted to short clips without hierarchical temporal assessment. Therefore, a unified benchmark for long-term video quality understanding, integrating perceptual fidelity, temporal coherence, and multimodal reasoning, is still lacking.

To bridge these gaps, we introduce \textbf{LongVQUBench}, a comprehensive benchmark for evaluating the LVQU of LVLMs. The benchmark contains over $1,200$ videos drawn from diverse sources, including films, documentaries, surveillance footage, egocentric recordings, and computer-generated content, spanning durations from a few minutes to nearly two hours. This diversity encompasses a wide range of perceptual distortions, such as lighting drift, scene transitions, codec artifacts, and generative distortions, that emerge over extended viewing periods. To systematically evaluate LVQU across increasing temporal scopes, LongVQUBench defines three hierarchical levels: \textbf{local event quality understanding~(LQU)}, \textbf{cross-event quality reasoning~(CQR)}, and \textbf{global quality understanding~(GQU)}. These levels progressively test a model’s capacity to identify localized distortions, integrate perceptual cues across events, and evaluate holistic perceptual integrity and temporal stability.

Beyond this hierarchical structure, LongVQUBench introduces a \textbf{needle distortion question-answering (NDQA)} paradigm, in which spatial or temporal artifacts of varying intensities are sparsely embedded throughout long videos. NDQA enables the analysis of fine-grained perceptual sensitivity and challenges LVLMs to reason beyond coarse semantic cues. We evaluate $14$ state-of-the-art LVLMs under zero-shot settings. The results show a consistent decline in performance as reasoning depth increases, revealing limitations in temporal localization, distortion attribution, and global quality reasoning. LongVQUBench establishes the first systematic benchmark for long-term video quality understanding, providing a principled framework to advance perceptual modeling, temporal reasoning, and multimodal integration toward human-level long-form video comprehension. 

Before delving into detail, we highlight our main contributions as follows:
\begin{itemize}[label=\sqbullet, leftmargin=*, itemsep=0pt, topsep=2pt]
\item A comprehensive benchmark, \textbf{LongVQUBench}, specifically designed to evaluate the long-term video quality understanding capability of LVLMs across diverse real-world content.
\item A hierarchical evaluation framework, encompassing local, cross-event, and global quality understanding, complemented by the needle distortion question-answering (NDQA) paradigm to probe fine-grained perceptual sensitivity.
\item A large-scale empirical study involving $14$ state-of-the-art LVLMs, which exposes fundamental limitations in temporal localization, perceptual attribution, and global reasoning across extended durations.
\end{itemize}

\section{Related Work}

\subsection{Large Vision-Language Models}
Large vision–language models (LVLMs) have significantly advanced multimodal understanding by aligning visual and textual modalities within unified generative frameworks. Foundational models such as CLIP~\cite{radford2021learning} and ALIGN~\cite{jia2021scaling} established scalable pretraining paradigms based on contrastive vision–language alignment, while BLIP~\cite{li2022blip} and BLIP-2~\cite{li2023blip2} introduced modular strategies that integrate pretrained language models with frozen vision encoders for efficient cross-modal learning. Instruction-tuned LVLMs, including Flamingo~\cite{alayrac2022flamingo}, GPT-5~\cite{openai2025gpt5}, LLaVA~\cite{liu2023llava}, InstructBLIP~\cite{dai2024instructblip}, and Qwen2.5-VL~\cite{xie2025qwen2_5vl}, have achieved strong generalization in vision-language reasoning, demonstrating remarkable progress across captioning, question answering, and grounding tasks.

Recent research has extended LVLMs toward multi-image and long-context video understanding. Models like Video-LLaVA~\cite{damon2024videollava}, LongVA~\cite{wu2024longvideobench}, Co-Instruct~\cite{wu2024towards}, and mPLUG-Owl3~\cite{ye2024mplugowl3} enhance temporal modeling by processing sequential frames or dynamic clips with interleaved text–video inputs. Similarly, LLaVA-Next~\cite{liu2024llavanext} and VideoChat~\cite{li2023videochat} improve long-sequence reasoning through efficient frame sampling, recurrent memory fusion, and temporal attention. Despite these advances, most LVLMs remain limited by short-context constraints and lack explicit mechanisms for modeling cumulative perceptual changes over extended durations. This limitation underscores the need for dedicated benchmarks such as \textbf{LongVQUBench}, which evaluate perceptual reasoning and temporal quality understanding in realistic long-form video settings.

\begin{table*}[t!]
\centering
\renewcommand{\arraystretch}{1.0}
\caption{
Comparison of \textbf{LongVQUBench} with existing benchmarks. 
Columns report the number of videos (\textbf{\#Vid.}), number of QA pairs (\textbf{\#QA}), average video duration in seconds (\textbf{Len.}), support for Multiple-choice questions~(\textbf{MCQ}) and open-ended questions, coverage of diverse genres (\textbf{Diverse Genres}), multi-duration evaluation (\textbf{Multi-Level}), and capability for video quality understanding~(\textbf{VQ Underst.}). 
The upper block lists general video understanding benchmarks, while the lower block focuses on video quality assessment benchmarks.
}
\label{tab:benchmark_comparison}

\begin{adjustbox}{max width=\textwidth}
\begin{tabular}{c>{\centering\arraybackslash}m{1.1cm}>{\centering\arraybackslash}m{1.2cm}>{\centering\arraybackslash}m{1.1cm}>{\centering\arraybackslash}m{1.2cm}>{\centering\arraybackslash}m{1.2cm}>{\centering\arraybackslash}m{1.3cm}>{\centering\arraybackslash}m{1.2cm}>{\centering\arraybackslash}m{1.4cm}}
\toprule
\textbf{Benchmarks} & \textbf{\#Vid.} & \textbf{\#QA} & \textbf{Len. (s)} & \textbf{MCQ} & \textbf{Open-ended} & \textbf{Diverse Genres} & \textbf{Multi-Level} & \textbf{VQ Underst.} \\
\midrule
Movie101~\cite{movie101-2023} & 101 & - & 6144 & \crossmark & \checkmark & \crossmark & \crossmark & \crossmark \\
EgoSchema~\cite{egoschema2023} & 5,063 & 5,063 & 180 & \checkmark & \crossmark & \crossmark & \crossmark & \crossmark \\
MovieChat-1K~\cite{song2024moviechat} & 1000 & 13K & 500 & \checkmark & \checkmark & \crossmark & \crossmark & \crossmark \\
Video-MME~\cite{videomme2024} & 900 & 2,700 & 1024 & \checkmark & \crossmark & \checkmark & \checkmark & \crossmark \\
LongVideoBench~\cite{wu2024longvideobench} & 3,763 & 6,678 & 473 & \checkmark & \crossmark & \checkmark & \checkmark & \crossmark \\
MLVU~\cite{zhao2025mlvu}  & 1,730 & 3,102 & 930 & \checkmark & \checkmark & \checkmark & \checkmark & \crossmark \\
\midrule
Q-Bench-Video~\cite{zhang2025qbench}  & 1,800 & 2,378 & 10 & \checkmark & \checkmark & \checkmark & \crossmark & \checkmark \\
\textbf{LongVQUBench}  & 1,200 & 1,500 & 742.2 & \checkmark & \checkmark & \checkmark & \checkmark & \checkmark \\
\bottomrule
\end{tabular}
\end{adjustbox}
\end{table*}

\subsection{Benchmarks for LVU}
Benchmarking long-term video understanding (LVU) is crucial for evaluating a model’s capacity to capture extended temporal dependencies, multi-event reasoning, and narrative coherence. Early datasets such as ActivityNet~\cite{caba2015activitynet}, Kinetics~\cite{kay2017kinetics}, and Something-Something~\cite{goyal2017something} primarily assess short clips and isolated actions. Subsequent works including Movie101~\cite{movie101-2023}, EgoSchema~\cite{egoschema2023}, and MovieChat-1K~\cite{song2024moviechat} extend evaluation to narrative and egocentric contexts, while LongVideoBench~\cite{wu2024longvideobench},  Video-MME~\cite{videomme2024}, and MLVU~\cite{zhao2025mlvu} advance multi-task and long-context reasoning across diverse domains. Q-Bench-Video~\cite{zhang2025qbench} further explores video quality understanding but remains limited to short sequences. In contrast, \textbf{LongVQUBench} introduces hierarchical, perceptually grounded evaluation across minute-to-hour videos, integrating temporal coherence, multi-level reasoning, and fine-grained perceptual quality assessment (see Table~\ref{tab:benchmark_comparison}).

\subsection{VQA Methods}

Video quality assessment~(VQA) aims to estimate human perceptual judgment of visual fidelity and temporal stability. Classical full-reference~(FR) methods, such as PSNR and SSIM~\cite{wang2004ssim}, model low-level signal fidelity, while perceptually motivated models like VMAF~\cite{li2016vmaf} and MOVIE~\cite{seshadrinathan2011motion}, integrate spatial and temporal features. No-reference~(NR) approaches, including NIQE~\cite{mittal2012niqe}, BRISQUE~\cite{mittal2012brisque}, and VIDEVAL~\cite{tu2021ugc}, estimate quality without reference images through handcrafted or learned statistical priors.
With deep learning, advanced FR and NR methods such as DeepVQA~\cite{kim2018deep}, C3DVQA~\cite{xu2020c3dvqa},  VSFA~\cite{li2019vsfa}, Fast-VQA~\cite{wu2022fast}, and DOVER~\cite{wu2023exploring} leverage perceptual feature representations to achieve stronger correlation with human opinion scores. Recent transformer-based~\cite{zhu2023learning,zhu2024video,wen2024modular} and multimodal approaches~\cite{jia2025vqa2,wang2025love,zhang2025vq} further enhance temporal modeling and generalization.
While these methods effectively predict short-term quality, they lack the reasoning ability and temporal context understanding required for long-form video analysis. \textbf{LongVQUBench} bridges this gap by integrating perceptual modeling with language-based reasoning to evaluate long-term video quality understanding in LVLMs.

\section{LongVQUBench}

This section details the design philosophy, data composition, and evaluation structure of \textbf{LongVQUBench}, a large-scale benchmark for assessing the long-term video quality ability of LVLMs. The benchmark is developed to jointly evaluate perceptual fidelity and temporal reasoning across extended durations, establishing a unified testbed for comprehensive analysis.

\subsection{Overview}
\begin{figure}[t]
    \centering
   \includegraphics[width=\linewidth, trim=0 0 0 0, clip]{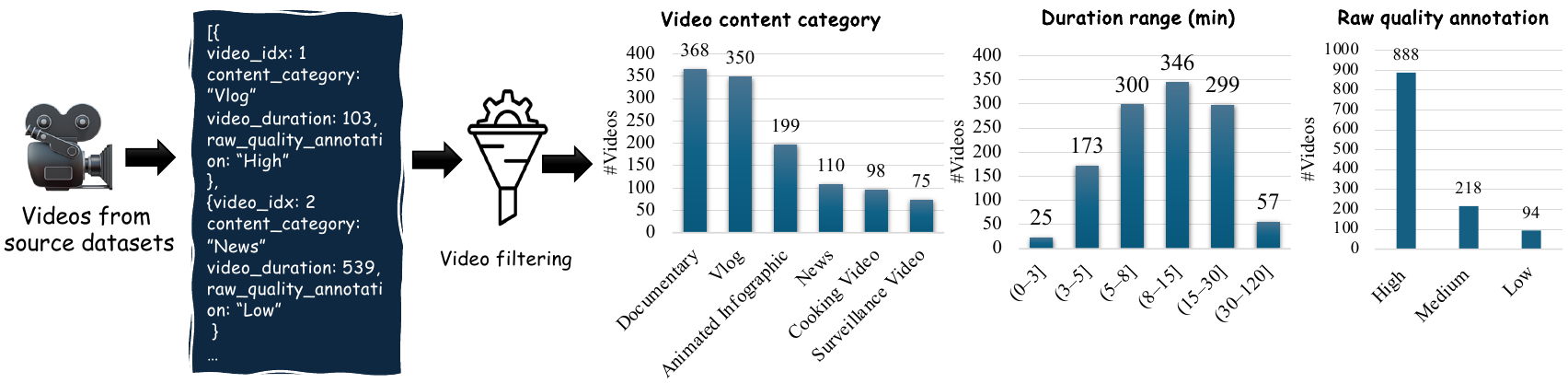}
    \caption{Long-term duration videos from LongVideoBench~\cite{wu2024longvideobench}, MLVU~\cite{zhao2025mlvu}, and LongVideoReason~\cite{chen2025scaling} are first aggregated and tagged, followed by a filtering and removal process to achieve the target distribution of LongVQUBench.}
    \label{fig:video_distribution}
\end{figure}

\textbf{LongVQUBench} is designed to examine how LVLMs perceive, reason, and explain video quality over long temporal horizons. Unlike existing short-clip datasets~\cite{zhang2025qbench}, it focuses on the gradual evolution of quality, the accumulation of perceptual degradation, and the reasoning dependencies among temporally distant events.
The benchmark is constructed with three design goals: 1) To cover a broad spectrum of long-form videos that reflect diverse real-world and synthetic scenarios; 2) To ensure representative coverage across both video duration and perceptual quality; and
3) To enable structured, hierarchical evaluation of temporal reasoning and fine-grained perceptual sensitivity.
An overview of the dataset composition is presented in Figure~\ref{fig:video_distribution}, which summarizes category distributions, duration statistics, and quality-level balance.

\subsection{Long-term Video Collection}

\textbf{LongVQUBench} contains \textbf{1,200 videos} sourced from publicly available datasets and open media collections, including LongVideoBench~\cite{wu2024longvideobench}, MLVU~\cite{zhao2025mlvu}, and LongVideoReason~\cite{chen2025scaling}. Each video spans a duration between several minutes and two hours, enabling analysis of temporal consistency and cross-event degradation in realistic viewing conditions.

\noindent\textbf{Diverse Video Sources.}
The videos cover multiple domains, including feature films, documentaries, surveillance footage, egocentric recordings, instructional videos, and computer-generated scenes. This diversity captures a wide range of motion patterns, editing styles, and semantic structures, ensuring the benchmark’s generality. Such variety also reflects the heterogeneous perceptual challenges faced by LVLMs when analyzing long-form content.

\noindent\textbf{Comprehensive Video Length.}
As shown in Figure~\ref{fig:video_distribution}, LongVQUBench includes a balanced distribution of video durations. Approximately one quarter of the dataset consists of short videos under 10 minutes, while the remainder spans medium (10 - 30 minutes) and long (30 - 120 minutes) durations. This coverage allows comprehensive analysis of performance trends as the temporal context increases.

\begin{figure}[t]
    \centering
   \includegraphics[width=\linewidth, trim=0 120 0 0, clip]{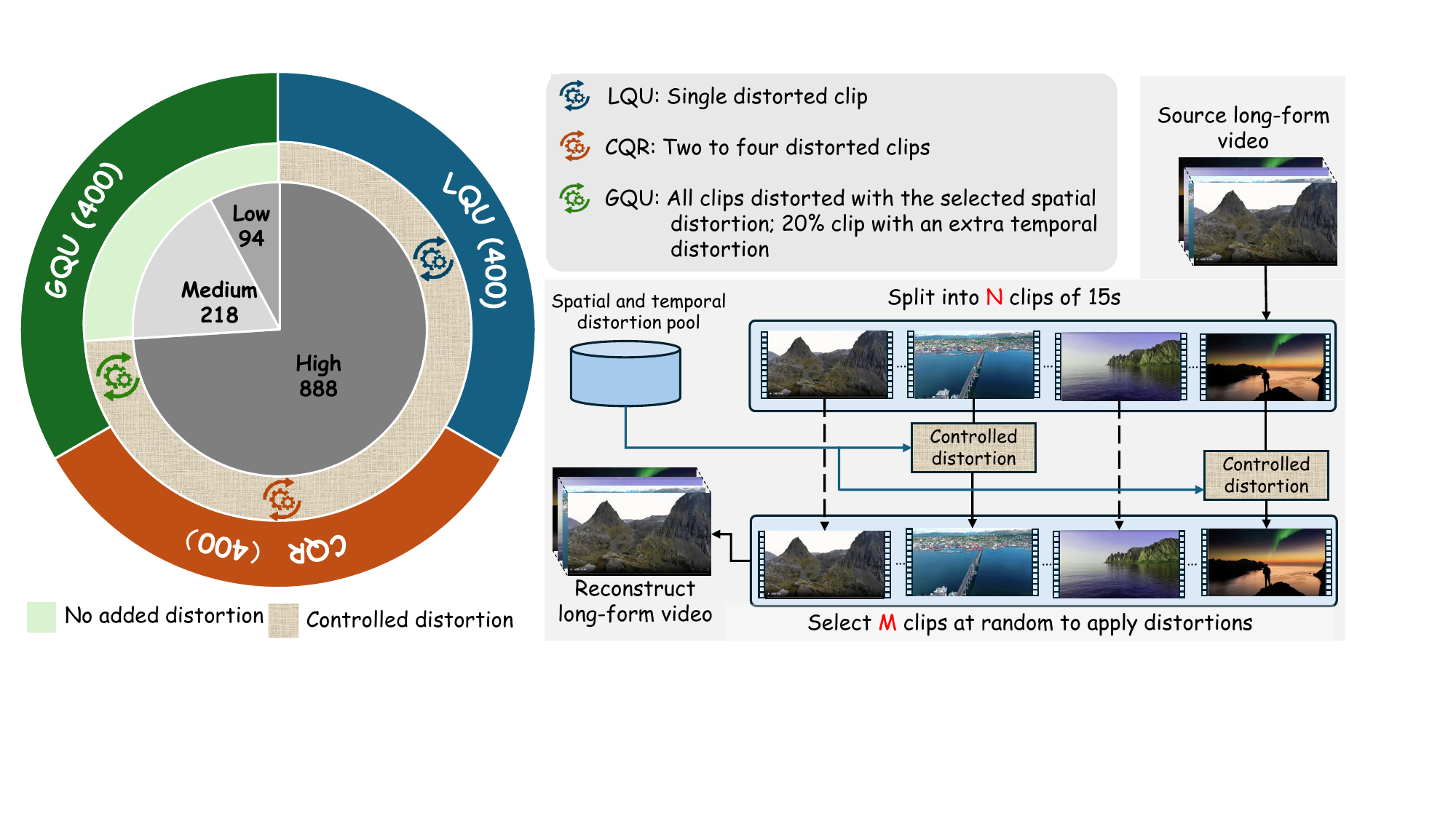}
    \caption{Left: Distribution of videos across hierarchical evaluation levels along with the proportion of samples subjected to controlled distortions.
Right: Illustration of the controlled distortion pipeline. High-quality videos are first segmented into 15-second clips. Controlled distortions are then applied according to predefined distortion pools and configurations (LQU, CQR, GQU). Finally, distorted clips are merged into full-length videos to enable distortion-aware question-answer generation.}
    \label{fig:controlled_distortions}
\end{figure}

\noindent\textbf{Comprehensive Quality Range.}
To ensure perceptual diversity, the dataset encompasses three quality levels: 
\textit{high}~(H), \textit{moderate}~(M), and \textit{low}~(L), as illustrated in Figure~\ref{fig:video_distribution}. 
These levels correspond to varying degrees of compression, lighting instability, motion jitter, and generative distortion.  The majority of videos remain of high perceptual quality (H = 888), complemented by moderate (M = 218) and low (L = 94) quality samples. 
This distribution is intentional: controlled distortions are systematically applied only to high-quality source videos, as shown in Figure~\ref{fig:controlled_distortions}.
Starting with clean visual content enables precise manipulation of distortion type and intensity, allowing us to generate distortion-aware question-answer pairs with reliable ground-truth alignment.

\subsection{Benchmark Construction}

\textbf{LongVQUBench} is constructed to systematically evaluate the perceptual reasoning capability of LVLMs across long-duration videos. Each video is paired with question-answer (QA) item(s) that assess how well models perceive, reason, and explain video quality under varying perceptual and temporal conditions. 
The benchmark integrates a hierarchical evaluation framework spanning local, cross-event, and global reasoning levels, complemented by a \textit{Needle Distortion Question-Answering (NDQA)} paradigm for fine-grained sensitivity assessment.

\noindent\textbf{Distortion Configuration.}
To simulate realistic degradation patterns, our benchmark incorporates a diverse set of \textit{spatial} and \textit{temporal} distortions that serve as the foundation for NDQA construction, as illustrated in Figure~\ref{fig:controlled_distortions}. The dataset includes $14$ spatial and $4$  temporal distortion types chosen for their relevance to common video production, transmission, and generative scenarios. Spatial distortions affect frame-level fidelity, while temporal distortions impact motion continuity and global temporal stability. Each distortion is implemented at three controlled intensity levels to ensure perceptual diversity. Detailed distortion types and parameter configurations are provided in the supplementary material.

\noindent\textbf{Needle Distortion Question-Answering (NDQA).} The NDQA paradigm employs the aforementioned distortions to probe models’ ability to detect and interpret degradations embedded within long videos. In this setting, spatial or temporal distortions of varying amplitudes are introduced without disrupting semantic content or narrative flow. Models are evaluated through two complementary question formats:
\begin{itemize}
    \item \textbf{Multiple-choice questions (MCQ)} include four types: \textit{Yes-or-No}, \textit{What}, \textit{Which}, and \textit{How}. The \textit{Yes-or-No} type examines the presence of perceptual degradations, such as detecting whether flicker or blur occurs within a segment. The \textit{What} type identifies the specific distortion category, while the \textit{How} type quantifies its perceptual strength or temporal extent. The \textit{Which} type requires comparative reasoning, prompting the model to determine which segment or event exhibits more severe degradation. Together, these question types jointly evaluate recognition accuracy, comparative judgment, and sensitivity to perceptual intensity.
    \item \textbf{Open-ended questions}, which require free-form reasoning, prompting models to describe the degradation’s nature, location, and perceptual impact.
\end{itemize}

This dual-format design unifies objective accuracy and interpretive evaluation, forming a balanced framework to measure both perceptual sensitivity and reasoning depth. 
Building on the NDQA paradigm, the subsequent three levels, \textit{Local Event Quality Understanding (LQU)}, \textit{Cross-Event Quality Reasoning (CQR)}, and \textit{Global Quality Understanding (GQU)}, extend the assessment toward progressively broader temporal and perceptual contexts.

\begin{figure}[t]
    \centering
   \includegraphics[width=\linewidth, trim=10 10 30 30, clip]{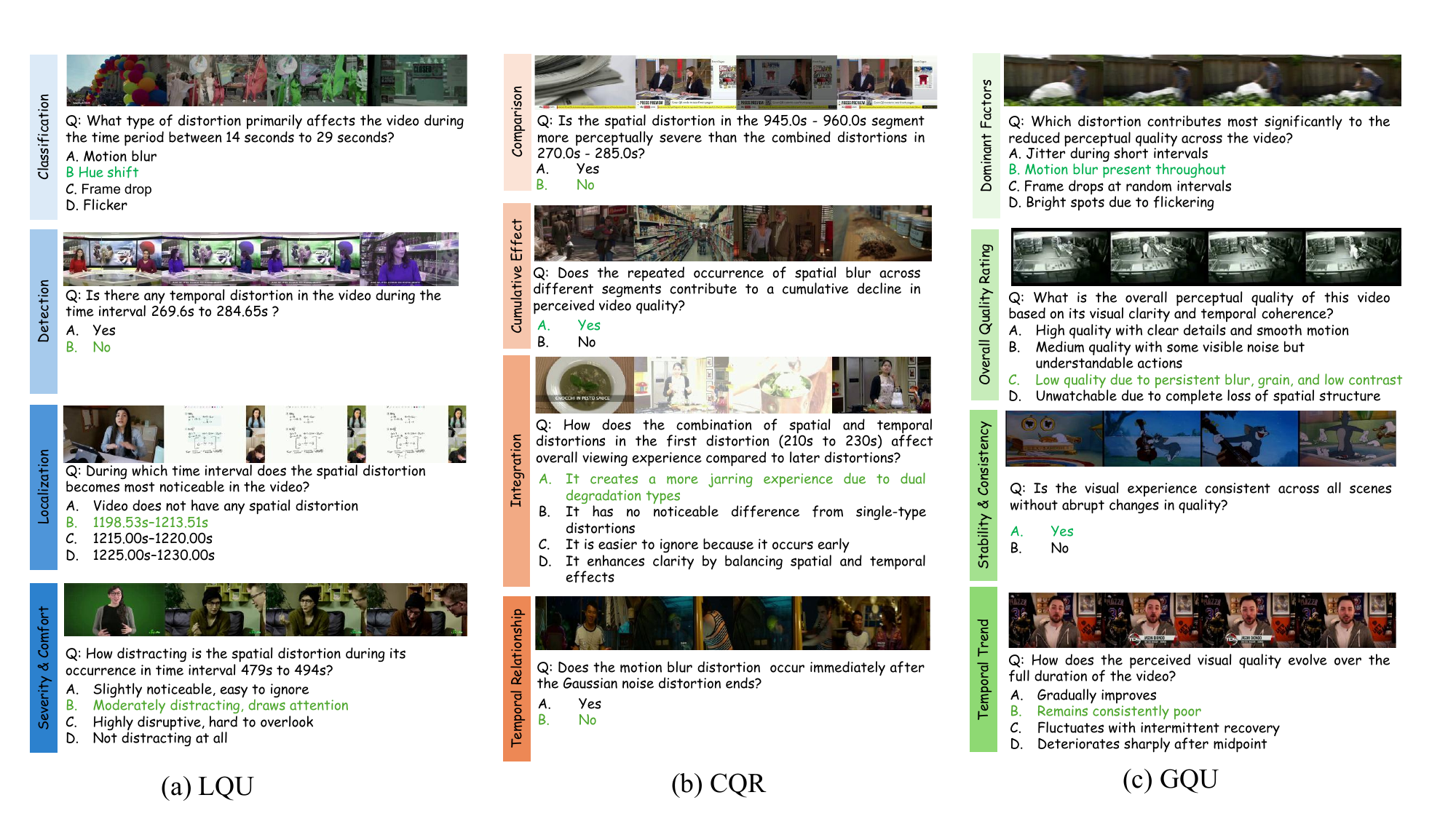}
    \caption{LongVQUBench features perceptual quality reasoning questions across multiple temporal scopes: (a) Local Event Quality Understanding (LQU) for analyzing localized distortions; (b) Cross-Event Quality Reasoning (CQR) for integrating multiple degraded events; and (c) Global Quality Understanding (GQU) for holistic perceptual evaluation over extended durations. }
    \label{fig:question_samples}
\end{figure}

\subsubsection{1) Local Event Quality Understanding (LQU):}

The LQU level evaluates a model’s ability to detect, classify, localize, and interpret a single, temporally bounded distortion event within a long video~\cite{gao2017tall,lin2023univtg}. Each event typically spans $5$ to $20$ seconds and reflects transient quality degradation phenomena that have become increasingly important in video quality understanding and analysis~\cite{zhang2025qbench,jia2025vqa2}, such as localized blur~\cite{liu2017quality,jia2025vqa2}, compression noise~\cite{zhang2022multi,zhang2025badiff}, luminance fluctuation~\cite{choi2018video,choi2015motion}, or flicker~\cite{choi2013visibility,choi2018video}, which can affect short-term perceptual comfort and visual attention~\cite{terzic2016methods,choi2015motion}.
LQU primarily tests the model’s short-term perceptual sensitivity and its capacity to link local distortions with subjective viewing discomfort.

\noindent\textbf{Question Design.}
Each LQU sample is associated with a question targeting one of five complementary perceptual dimensions:
\begin{itemize}
    \item \textit{Detection} determines whether a perceptual distortion exists within a given temporal segment.
    \item \textit{Localization} identifies when the degradation occurs in the video timeline.
    \item \textit{Classification} specifies the distortion category, such as blur, flicker, or color shift.
    \item \textit{Severity and Comfort Assessment} estimates how intense, distracting, or perceptually disturbing the distortion appears.
    \item \textit{Open Reasoning} explains why the observed artifact negatively affects perceived quality, focusing on aspects like motion inconsistency or visual discomfort.
\end{itemize}

We have shown the sampled questions based on the above design in Figure~\ref{fig:question_samples}(a). This progressive questioning structure encourages models to move beyond binary judgment toward fine-grained perceptual reasoning. Each QA pair is manually validated to ensure visual clarity, temporal precision, and interpretive consistency, enabling objective evaluation of local perceptual sensitivity.

\subsubsection{2) Cross-Event Quality Reasoning (CQR):}

The CQR level examines a model’s ability to compare, associate, and integrate multiple distortion events distributed across a long video. Unlike LQU, which focuses on short-term perceptual sensitivity, CQR targets reasoning across extended temporal spans where multiple degradations may occur sequentially or intermittently. This level evaluates whether a model can assess the relative severity of distortions, capture their temporal relationships, and infer how their interactions influence overall perceptual comfort.

\noindent\textbf{Question Design.}
Each CQR instance is designed to measure the model's capability to conduct multi-event reasoning across one of five complementary dimensions:
\begin{itemize}
    \item \textit{Comparison} identifies which segment or scene exhibits stronger or more disturbing degradations.
    \item \textit{Cumulative Effect} assesses how the accumulation of multiple artifacts influences perceptual stability or viewer fatigue.
    \item \textit{Integration} determines whether the model can synthesize perceptual evidence from multiple segments to form a consistent overall judgment.
    \item \textit{Temporal Relation} evaluates whether the distortions are temporally correlated, clustered, or independently distributed.
    \item \textit{Open Reasoning} requires the model to explain how different distortions interact over time, emphasizing contextual reasoning beyond local perception.
\end{itemize}

Together, these dimensions position CQR as a bridge between local perceptual analysis and holistic quality interpretation. Each annotated QA pair is manually verified to ensure the spatial-temporal correspondence of events and to prevent content bias between compared segments. Several sampled questions are shown in Figure~\ref{fig:question_samples}(b). This structured formulation provides a controlled setting for testing the model’s ability to reason across temporal dependencies and accumulated degradations.

\subsubsection{3) Global Quality Understanding (GQU):}

The GQU level evaluates a model’s ability to synthesize an overall perceptual judgment across an entire long video, typically spanning from one minute to two hours. It requires reasoning about temporal trends, cumulative degradations, and perceptual stability over prolonged viewing. In contrast to LQU and CQR, which focus on local distortions or multi-event relations, GQU emphasizes holistic temporal integration, tracking how perceptual quality evolves, stabilizes, or deteriorates over time and how this evolution affects the final perceptual judgment. 

\noindent\textbf{Question Design.}
Each GQU instance aims to measure long-term perceptual coherence through one of five complementary dimensions:
\begin{itemize}
    \item \textit{Stability Evaluation} assesses the consistency of viewing experience, capturing long-term fluctuations and viewer fatigue.
    \item \textit{Dominant Factor Identification} determines the principal degradation type or event that most strongly influences the overall judgment.
    \item \textit{Trend Assessment} identifies whether the perceived quality improves, remains stable, or degrades as the video progresses.
    \item \textit{Overall Evaluation} estimates the global perceptual quality of entire video, integrating spatial fidelity, temporal smoothness, and aesthetic appeal.
    \item \textit{Open Reasoning} requires the model to explain why the overall perception aligns with its given judgment, articulating the temporal and perceptual evidence that supports its decision.
\end{itemize}

Representative questions generated according to the above design are shown in Figure~\ref{fig:question_samples}(c). This level bridges perceptual aggregation and interpretive reasoning, allowing the evaluation of whether LVLMs can move from event-based assessment to globally consistent quality judgments. All QA items are validated through expert review to ensure reliable temporal coverage and consistent interpretability, thereby establishing a foundation for analyzing holistic perceptual reasoning in long-duration videos.

\subsection{Questions \& Answers Annotation}

All QA pairs in \textbf{LongVQUBench} are annotated through a controlled two-stage process to ensure temporal accuracy, semantic clarity, and perceptual consistency. In the first stage, QA items are constructed based on the reasoning dimensions of each level: LQU, CQR, and GQU, by identifying distortion events, marking their temporal boundaries, and formulating questions that probe detection, comparison, and reasoning. In the second stage, each QA item is independently reviewed by multiple experts, with disagreements resolved through consensus to ensure annotation reliability. Both multiple-choice and open-ended formats are adopted: the former targets objective recognition and localization, while the latter evaluates explanatory reasoning, assisted by GPT-based scoring for relevance and completeness~\cite{zhang2025qbench, zhao2025mlvu}. This rigorous yet scalable annotation protocol guarantees consistency across perceptual levels and establishes a robust foundation for evaluating long-term video quality understanding. Further details of the annotation procedure are provided in the supplementary material.

\section{Results on LongVQUBench}

In this section, we first describe the experimental settings, including the participating LVLMs, the evaluation protocol, and the dataset split. We then present quantitative results and analyze the performance of current LVLMs on long-term video quality understanding. More experimental results can be found in the supplementary material.

\subsection{Experimental Settings}

\noindent\textbf{Benchmark LVLMs.}
We evaluate a total of \textbf{14 LVLMs}, including \textit{3 proprietary models}: GPT-5~\cite{openai2025gpt5}, Gemini~3~\cite{gemini3_google_2025}, and Qwen-VL-Max~\cite{bai2023qwen}; \textit{7 open-source models}: LLaVA-NeXT-Video~\cite{liu2024llavanext}, ShareGPT4Video~\cite{chen2024sharegpt4video}, Qwen3-VL~\cite{bai2025qwen3}, MovieChat~\cite{song2024moviechat}, LLaVA-Video~\cite{zhang2024llava}, VQA$^2$~\cite{jia2025vqa2}, and Long-RL~\cite{chen2025scaling}; and \textit{4 agentic LVLMs}: VideoAgent~\cite{VideoAgent}, VideoExplorer~\cite{yuan2025videoexplorer}, LongVT~\cite{yang2025longvt}, and DeepVideoDiscovery~\cite{zhang2025deep}. 
Notably, VQA$^2$ is specifically designed for video quality understanding. 
Together, these models cover a diverse set of architectures, training paradigms, and reasoning mechanisms, enabling a comprehensive evaluation of current LVLM capabilities for long-term video quality understanding. More details of these LVLMs can be found in the supplementary material.

\noindent
\textbf{Evaluation Protocol.}
We evaluate LVLMs using a frame-sampling-based inference pipeline designed for long-duration videos. 
Given a video $\mathcal{V}$ with duration $T$, we uniformly sample $n$ frames at a fixed frame rate (FPS). 
The sampled frame sequence $\{f_1, f_2, \dots, f_n\}$ is ordered chronologically and provided to the LVLM within a single prompt. 
The prompt explicitly specifies: 
(i) the total video duration, 
(ii) the sampling frame rate (FPS), 
(iii) the total number of sampled frames, and 
(iv) a strict output format constraint. 
The model is instructed to act as an expert in video quality analysis and must select \emph{exactly one} answer from the provided candidate options. 
Each question is evaluated in a single forward pass without iterative interaction.

\noindent
\textbf{Dataset and Split.}
Experiments are conducted on the \textbf{LongVQUBench} dataset, which evaluates long-duration video quality understanding across three hierarchical levels: LQU, CQR, and GQU, each comprising four question-design dimensions. 
To ensure balanced evaluation across all dimensions, we adopt a \textbf{stratified 40\% validation / 60\% test split}. 
The split is performed independently within each question dimension to preserve the proportional distribution of samples. 
The validation set is used to determine the optimal number of sampled frames ($\#frames$). 
Once the sampling configuration is selected, it is fixed and applied to the held-out test set for final evaluation.

\subsection{Main Results}

\noindent
\textbf{Validation Results.}
We first analyze model performance on the $40\%$ validation subset under different frame sampling budgets. 
Table~\ref{tab:val_results_num} reports results for open-source and proprietary LVLMs when the number of uniformly sampled frames is capped at 1 FPS. 
Several observations can be drawn.

\textit{1) Increasing frame count yields limited gains.} 
For most models, increasing the number of sampled frames does not consistently improve performance. 
Large proprietary models such as GPT-5 and Gemini-3 benefit from moderate increases in frame count, with GPT-5 achieving its best performance at 256 frames (74.1 overall) and Gemini-3 peaking at 128 frames (68.9 overall). 
However, the improvements quickly saturate beyond moderate sampling budgets. 
In contrast, several video-specialized open-source models achieve their best performance with relatively small inputs, e.g., VQA$^2$ performs best with only 8 frames (59.4 overall), and Qwen3-VL peaks at 64 frames (63.1 overall). 
These results indicate diminishing returns from increasing temporal coverage.

\begin{table}[t]
\centering
\caption{Results on the LongVQUBench {\tt val} subset for the long video quality perception ability of LVLMS according to the number of max frames (capped at 1 FPS).}

\label{tab:val_results_num}
\renewcommand{\arraystretch}{1.05}
\resizebox{\linewidth}{!}{%
\begin{tabular}{c|c|cccc|c|c|cccc}
\toprule
\rowcolor{gray!20}
\textbf{Model} &
  \textbf{\#frames} &
  \textbf{LQU} &
  \textbf{CQR} &
  \textbf{GQU} &
  \textbf{Total} &
  \textbf{Model} &
  \textbf{\#frames} &
  \textbf{LQU} &
  \textbf{CQR} &
  \textbf{GQU} &
  \textbf{Total} \\
\multirow{5}{*}{GPT-5} &
  8 &
  72.4 &
  77.8 &
  61.2 &
  70.5 &
  \multirow{5}{*}{LLaVA-Video} &
  \textbf{8} &
  \textbf{61.5} &
  \textbf{64.5} &
  \textbf{48.5} &
  \textbf{58.2} \\
 &
  32 &
  71.5 &
  78.5 &
  63.0 &
  71.0 &
   &
  16 &
  60.2 &
  64.0 &
  47.5 &
  57.2 \\
 &
  64 &
  74.0 &
  80.2 &
  64.5 &
  72.9 &
   &
  30 &
  57.5 &
  62.6 &
  47.0 &
  55.7 \\
 &
  128 &
  75.5 &
  81.6 &
  65.0 &
  74.0 &
   &
  128 &
  N/A &
  N/A &
  N/A &
  N/A \\
 &
  \textbf{256} &
  \textbf{75.2} &
  \textbf{81.2} &
  \textbf{65.8} &
  \textbf{74.1} &
   &
  256 &
  N/A &
  N/A &
  N/A &
  N/A \\
  \midrule
\multirow{5}{*}{Gemini-3} &
  8 &
  71.2 &
  73.6 &
  59.6 &
  68.1 &
  \multirow{5}{*}{VQA$^2$} &
  \textbf{8} &
  \textbf{63.2} &
  \textbf{65.5} &
  \textbf{49.5} &
  \textbf{59.4} \\
 &
  32 &
  67.2 &
  75.0 &
  56.2 &
  66.1 &
   &
  32 &
  59.0 &
  63.5 &
  47.5 &
  56.7 \\
 &
  64 &
  67.0 &
  74.2 &
  58.5 &
  66.6 &
   &
  64 &
  59.5 &
  64.0 &
  52.0 &
  58.5 \\
 &
  \textbf{128} &
  \textbf{71.0} &
  \textbf{76.5} &
  \textbf{59.2} &
  \textbf{68.9} &
   &
  128 &
  58.0 &
  63.0 &
  50.5 &
  57.2 \\
 &
  256 &
  68.8 &
  75.6 &
  57.6 &
  67.3 &
   &
  256 &
  55.5 &
  61.0 &
  48.0 &
  54.8 \\
  \midrule
\multirow{5}{*}{Qwen-VL-Max} &
  8 &
  68.2 &
  70.4 &
  48.2 &
  62.3 &
  \multirow{5}{*}{Long-RL} &
  8 &
  62.5 &
  66.7 &
  50.0 &
  59.7 \\
 &
  32 &
  64.5 &
  70.8 &
  55.0 &
  63.4 &
   &
  32 &
  61.5 &
  67.2 &
  50.0 &
  59.6 \\
 &
  \textbf{64} &
  \textbf{64.5} &
  \textbf{73.5} &
  \textbf{54.5} &
  \textbf{64.2} &
   &
  \textbf{64} &
  \textbf{61.7} &
  \textbf{67.0} &
  \textbf{55.0} &
  \textbf{61.2} \\
 &
  128 &
  62.0 &
  67.5 &
  53.0 &
  60.8 &
   &
  128 &
  60.5 &
  66.8 &
  52.0 &
  59.8 \\
 &
  250 &
  61.0 &
  65.8 &
  51.2 &
  59.3 &
   &
  256 &
  57.2 &
  66.0 &
  49.2 &
  57.5 \\
  \midrule
\multirow{5}{*}{LLaVA-NeXT-Video} &
  \textbf{8} &
  \textbf{46.6} &
  \textbf{75.2} &
  \textbf{50.2} &
  \textbf{57.3} &
  \multirow{5}{*}{ShareGPT4Video} &
  8 &
  30.5 &
  37.2 &
  20.0 &
  29.2 \\
 &
  16 &
  45.8 &
  68.2 &
  42.0 &
  52.0 &
   &
  \textbf{16} &
  \textbf{33.5} &
  \textbf{35.0} &
  \textbf{20.5} &
  \textbf{29.7} \\
 &
  30 &
  45.8 &
  67.8 &
  38.7 &
  50.8 &
   &
  64 &
  32.2 &
  34.5 &
  20.0 &
  28.9 \\
 &
  64 &
  N/A &
  N/A &
  N/A &
  N/A &
   &
  128 &
  N/A &
  N/A &
  N/A &
  N/A \\
 &
  128 &
  N/A &
  N/A &
  N/A &
  N/A &
   &
  256 &
  N/A &
  N/A &
  N/A &
  N/A \\
  \midrule
\multirow{5}{*}{Qwen3-VL} &
  8 &
  51.8 &
  59.6 &
  49.5 &
  53.6 &
  \multirow{5}{*}{MovieChat} &
  8 &
  36.5 &
  38.7 &
  21.2 &
  32.1 \\
 &
  32 &
  54.5 &
  60.0 &
  50.0 &
  54.8 &
   &
  32 &
  37.5 &
  42.5 &
  26.0 &
  35.3 \\
 &
  \textbf{64} &
  \textbf{54.0} &
  \textbf{75.0} &
  \textbf{60.4} &
  \textbf{63.1} &
   &
  \textbf{64} &
  \textbf{35.0} &
  \textbf{43.6} &
  \textbf{30.0} &
  \textbf{36.2} \\
 &
  128 &
  52.0 &
  58.2 &
  51.0 &
  53.7 &
   &
  128 &
  34.2 &
  42.0 &
  31.5 &
  35.9 \\
 &
  256 &
  49.5 &
  56.5 &
  49.5 &
  51.8 &
   &
  256 &
  33.0 &
  41.5 &
  30.0 &
  34.8\\
  \bottomrule
\end{tabular}}
\end{table}

\begin{table}[t]
\centering
\setlength{\tabcolsep}{5pt}
\caption{Results on the LongVQUBench {\tt val} subset for the long video quality perception ability of Agentic LVLMs.
Keyframes are adaptively selected - details provided in the supplementary material.}
\label{tab:val_results_agents}
\resizebox{0.6\linewidth}{!}{%
\begin{tabular}{ccccc}
\toprule

\textbf{Agent}     & \textbf{LQU} & \textbf{CQR} & \textbf{GQU} & \textbf{Overall} \\
\midrule
VideoAgent         & 34.2         & 42.5         & 30.7         & 35.8           \\
VideoExplorer      & \underline{44.5}         & 58.5         & 54.0         & 52.3           \\
LongVT             & 43.7         & \underline{65.0}         & \underline{60.0}         & \underline{56.2}           \\
DeepVideoDiscovery & \textbf{69.2}         &\textbf{ 82.7}         & \textbf{63.2 }        & \textbf{71.7} \\       \bottomrule
\end{tabular}}
\end{table}

\textit{2) Global quality reasoning remains challenging.} 
Across nearly all models, performance follows a consistent pattern where LQU and CQR scores exceed GQU scores. 
For example, GPT-5 achieves 81.2 on CQR but only 65.8 on GQU, while Gemini-3 obtains 76.5 on CQR versus 59.2 on GQU. 
This gap suggests that current LVLMs are more capable of detecting localized or cross-event distortions than synthesizing holistic quality judgments over long temporal horizons.

Table~\ref{tab:val_results_agents} presents validation results for agentic LVLMs that employ adaptive keyframe selection strategies. 
Compared with uniform frame sampling, the effectiveness of adaptive selection varies substantially across methods. 
DeepVideoDiscovery significantly outperforms other agentic approaches, achieving 71.7 overall accuracy with particularly strong performance on LQU (69.2) and CQR (82.7), approaching the level of leading proprietary LVLMs. 
In contrast, simpler agent-based systems such as VideoAgent and VideoExplorer achieve substantially lower scores. 
These results indicate that while adaptive frame selection can be beneficial, its effectiveness strongly depends on the quality of the exploration and reasoning strategy used to identify informative frames.

\begin{table}[t]
\caption{{\tt Test} leaderboard of LongVQUBench across 14 LVLMs, organized by hierarchical quality understanding levels and MCQ question dimensions. Abbreviations denote the following terms: \#F: Number of Frame; D: Detection; L: Localization; C: Classification; SA: Severity \& Comfort Assessment; CMP: Comparison; CE: Cumulative Effect; I: Integration; TR: Temporal Relation; SE: Stability Evaluation; DFI: Dominant Factor Identification; TA: Trend Assessment; OE: Overall Evaluation.}
\label{tab:test_results}
\vspace{-5pt}
\renewcommand{\arraystretch}{1.15}
\setlength{\tabcolsep}{4.8pt} % widen column padding (default ~6pt; adjust 4.5--6.5)

\resizebox{\linewidth}{!}{
\begin{tabular}{c|c|cccc@{\hskip 6pt}|cccc@{\hskip 6pt}|cccc|c}
\toprule
\multirow{2}{*}{Model} & \multirow{2}{*}{\#F}
& \multicolumn{4}{c|}{\textbf{LQU}}
& \multicolumn{4}{c|}{\textbf{CQR}}
& \multicolumn{4}{c|}{\textbf{GQU}}
& \multirow{2}{*}{\textbf{Overall}} \\
\cline{3-14}
& & D & L & C & SA
  & CMP & CE & I & TR
  & SE & DFI & TA & OE
  & \\
\hline
\rowcolor{gray!10}
\multicolumn{15}{c}{\textit{Closed-source LVLMs}} \\
\hline
GPT-5        & 256 & \textbf{84.6} & 71.5 & \textbf{56.5} & \underline{49.0} &\textbf{ 71.5} & \textbf{87.6} & \underline{86.5} & \underline{83.0 }& \textbf{68.5} & \textbf{65.0} & \textbf{48.5} & \textbf{61.5} & \textbf{69.5} \\
Gemini-3     & 128 & 76.5 & 68.0 & 54.0 & 48.0 & \underline{70.0} & \underline{85.0} & {82.0} & 80.0 & \underline{67.0} & 63.0 & 47.5 & \underline{60.0} & \underline{66.8} \\
Qwen-VL-Max  & 64  & 70.0 & 65.5 & 51.0 & 44.0 & 68.0 & 84.0 & 80.0 & 77.0 & 65.0 & 60.0 & 45.5 & 58.0 & 64.0 \\
\hline
\rowcolor{gray!10}
\multicolumn{15}{c}{\textit{Video LVLMs}} \\
\hline
LLaVA-NeXT-Video & 8  & 35.0 & 58.7 & 46.7 & 38.3 & 61.7 & 70.3 & 74.7 & 73.3 & 58.3 & 50.3 & 58.3 & 51.7 & 56.4 \\
ShareGPT4Video   & 16 & 26.5 & 34.6 & 28.0 & 32.6 & 24.6 & 35.2 & 34.0 & 33.0 & 23.6 & 28.0 & 23.5 & 26.0 & 29.1 \\
Qwen3-VL         & 64 & 46.5 & 62.5 & 47.0 & 46.2 & 65.0 & 78.2 & 75.0 & 72.1 & 64.0 & 60.0 & \textbf{48.5} & 58.5 & 60.3 \\
MovieChat        & 64 & 25.2 & 41.8 & 30.0 & 23.0 & 36.5 & 43.6 & 41.2 & 39.3 & 44.0 & 40.0 & 25.0 & 38.0 & 35.6 \\
LLaVA-Video      & 8  & 30.5 & 57.0 & 43.0 & 36.0 & 66.0 & 76.0 & 72.0 & 68.0 & 59.0 & 54.0 & 39.0 & 53.0 & 54.5 \\
VQA$^2$          & 8  & 62.5 & 53.6 & 48.0 & 41.5 & 60.5 & 49.0 & 75.0 & 72.0 & 48.2 & 59.0 & 44.0 & 37.5 & 54.2 \\
Long-RL          & 64 & 73.3 & \underline{74.3} & \underline{48.3} & 51.7 & 38.3 & 53.3 & \textbf{93.3} & \textbf{83.3} & 53.3 & \underline{63.3} & 48.3 & 38.3 & 59.9 \\
\hline
\rowcolor{gray!10}
\multicolumn{15}{c}{\textit{Agentic LVLMs}} \\
\hline
VideoAgent         & -- & 25.0 & 30.5 & 30.5 & 28.5 & 29.5 & 35.0 & 47.5 & 48.0 & 33.5 & 40.6 & 33.0 & 40.6 & 35.2 \\
VideoExplorer      & -- & 33.5 & 49.0 & 35.0 & 40.2 & 52.0 & 57.0 & 63.0 & 60.2 & 49.5 & 44.0 & 42.5 & 43.5 & 47.5 \\
LongVT             & -- & 37.5 & 62.0 & 55.0 & 46.5 & 43.0 & 56.0 & 53.0 & 50.0 & 41.0 & 37.5 & 27.0 & 37.0 & 45.5 \\
DeepVideoDiscovery & -- & \underline{77.3} & \textbf{83.3} & \underline{56.3} & \textbf{59.7} & 66.3 & 67.3 & 81.3 & 73.3 & 63.3 & 61.3 & 46.3 & 56.3 & 66.0 \\
\bottomrule
\end{tabular}
}
\end{table}

\begin{table}
\centering
\caption{
{\tt Test} leaderboard of relevance (R) and completeness (C) scores (\%) for open-ended questions across hierarchical levels - LQU, CQR, and GQU.}
\resizebox{0.85\linewidth}{!}{%
\begin{tabular}{c|cccc|cccc}
\toprule
\textbf{Model} &
  \textbf{LQU$_R$} &
  \textbf{CQR$_R$} &
  \textbf{GQU$_R$} &
  \textbf{Overall$_R$} &
  \textbf{LQU$_C$} &
  \textbf{CQR$_C$} &
  \textbf{GQU$_C$} &
  \textbf{Overall$_C$} \\
  \midrule
\multicolumn{9}{c}{\cellcolor[HTML]{E7E6E6}Closed-Source LVLMs} \\
\midrule
GPT-5             & \textbf{88.2} & \textbf{89.4} & \textbf{89.1} & \textbf{88.9} & \textbf{45.2} & \textbf{47.8} & \textbf{44.3} & \textbf{45.8} \\
Gemini-3           & 86.3 & 86.1 & 85.6 & 86.0   & 38.3 & 42.3 & 38.5 & 39.7  \\
Qwen-VL-Max        & 83.4 & 82.7 & 84.9 & 83.7 & 36.9 & 40.3 & 37.2 & 38.1  \\
\midrule
\multicolumn{9}{c}{\cellcolor[HTML]{E7E6E6}Open-Source LVLMs} \\
\midrule

LLaVA-NeXT-Video   & 74.1 & 78.3 & 76.6 & 76.3 & 36.2 & 39.5 & 37.5 & 37.7  \\
ShareGPT4Video     & 71.5 & 76.3 & 75.2 & 74.3 & 35.0   & 38.2 & 36.9 & 36.7  \\
Qwen3-VL           & 77.2 & 81.5 & 79.3 & 79.3 & 37.9 & 40.8 & 38.8 & 39.2  \\
MovieChat          & 72.4 & 77.5 & 74.6 & 74.8 & 35.4 & 39.1 & 36.3 & 36.9  \\
LLaVA-Video        & 73.5 & 78.7 & 75.3 & 75.8 & 36.0   & 39.4 & 36.8 & 37.4  \\
VQA$^2$            & 73.0   & 78.1 & 74.9 & 75.3 & 35.6 & 39.4 & 36.6 & 37.2  \\
Long-RL            & 76.5 & 81.0   & 77.5 & 78.3 & 37.4 & 40.8 & 37.8 & 38.7  \\
\midrule 
\multicolumn{9}{c}{\cellcolor[HTML]{E7E6E6}Agentic LVLMs}  \\  
\midrule
VideoAgent         & 72.0   & 74.8 & 74.2 & 73.7 & 35.1 & 38.6 & 36.4 & 36.70 \\
VideoExplorer      & 76.3 & 78.2 & 77.5 & 77.3 & 37.4 & 40.7 & 37.9 & 38.67 \\
LongVT             & 75.2 & 77.4 & 76.4 & 76.3 & 36.8 & 40.6 & 37.3 & 38.23 \\
DeepVideoDiscovery & 80.3 & 82.6 & 81.1 & 81.3 & 39.4 & 43.0   & 39.6 & 40.67\\
\bottomrule
\end{tabular}
}
\label{tab:oe_relevance_completeness}
\end{table}

\begin{figure}[t]
    \centering
   \includegraphics[width=\linewidth, trim=0 155 10 10, clip]{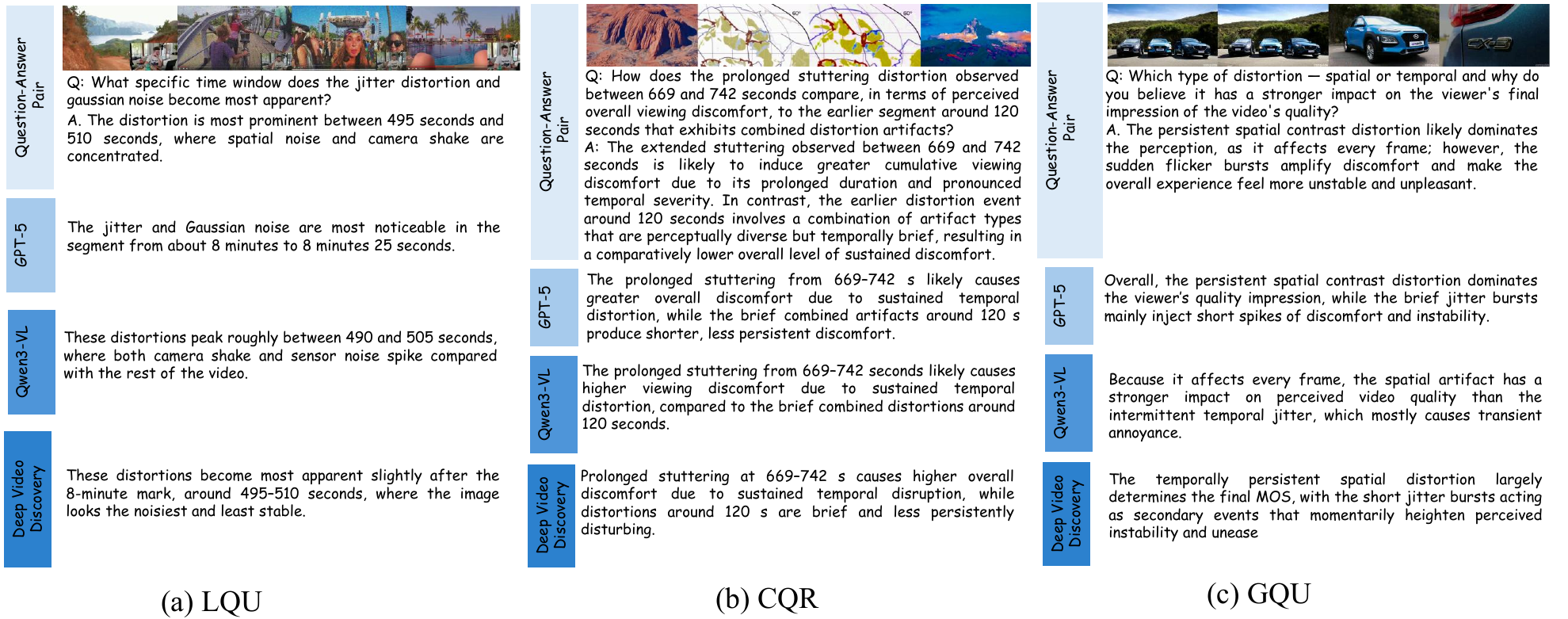}
    \caption{Open-ended evaluation: Example questions from LongVQUBench across three hierarchical categories (LQU, CQR and GQU), with corresponding answers from each of best-performing closed-source LVLM (GPT-5~\cite{openai2025gpt5}), open-source LVLM (Qwen3-VL~\cite{bai2025qwen3}), and agentic LVLM (DeepVideoDiscovery~\cite{zhang2025deep}).}
    \label{fig:oe_eval}
\end{figure}

\noindent
\textbf{Test Results.}
Using the frame configurations selected on the validation set, we report the final accuracy on the $60\%$ held-out test set in Table~\ref{tab:test_results} for multiple-choice questions. The Figure~\ref{fig:oe_eval} shows open-ended question-answer pair in dataset and answers from each of best-performing closed-source LVLM, open-source LVLM, and agentic LVLM. Table~\ref{tab:oe_relevance_completeness} reports the relevance and completeness scores of open-ended question answers across the hierarchical levels LQU, CQR, and GQU. The details of the relevance and completeness score of the GPT-based prompt are available in supplementary material.
All models are evaluated under a single-pass inference protocol without additional adaptation. 
The leaderboard provides fine-grained results across hierarchical quality understanding levels, including LQU, CQR, and GQU. 
Several key observations emerge.

\textit{1) Proprietary LVLMs achieve the strongest overall performance.}
Closed-source models dominate the leaderboard, with GPT-5 achieving the best overall accuracy (69.5), followed by Gemini-3 (66.8) and Qwen-VL-Max (64.0). 
These models show consistently strong performance across most reasoning dimensions, particularly in cross-event reasoning tasks.

\textit{2) Local event understanding is relatively tractable.}
LQU tasks achieve the high accuracy compared to GQU across the multiple-choice questions for open-sourced and closed-sourced LVLMs. 
While proprietary models perform strongly overall, some open-source and agentic systems also demonstrate competitive performance in specific dimensions, such as DeepVideoDiscovery achieving the highest localization accuracy (83.3).

\textit{3) Cross-event reasoning highlights the importance of temporal integration.}
CQR results show substantial variation across models. 
GPT-5 achieves strong performance in cumulative effect and comparison, while Long-RL achieves the best integration score (93.3), indicating that effective temporal aggregation is crucial for modeling interactions among multiple distortion events.

\textit{4) Global quality understanding remains the most challenging level.}
A consistent performance drop from LQU and CQR to GQU is observed across nearly all models, highlighting the difficulty of synthesizing long-term perceptual evidence into a coherent global quality judgment.

\textit{5) Agentic LVLMs show promising but uneven performance.}
Among agentic approaches, DeepVideoDiscovery achieves competitive performance (66.0 overall), approaching proprietary models and outperforming most LVLMs. Simpler agentic systems lag, showing effective frame exploration and reasoning are critical for agent-based long-video quality understanding.

\textit{6) Completeness remains a key challenge despite high relevance.}
Open-ended question relevance scores remain consistently high across all models, with closed-source models achieving the strongest performance (GPT-5: 88.9\%), indicating that most responses effectively address the given questions. However, completeness scores are substantially lower across all tiers, with even the best-performing model, GPT-5, reaching only 45.8\%. This reflects the inherent difficulty of producing comprehensive answers to open-ended quality assessment questions.

\section{Conclusion}

We introduce \textbf{LongVQUBench}, the first large-scale benchmark designed to evaluate long-term video quality understanding in LVLMs. Unlike existing short-clip quality assessment datasets or long-video semantic benchmarks, our benchmark integrates perceptual fidelity, temporal coherence, and reasoning across extended durations through a hierarchical evaluation framework together with the NDQA paradigm for fine-grained perceptual probing.
Extensive experiments on 14 state-of-the-art LVLMs reveal that model performance consistently degrades as temporal span and reasoning complexity increase, and that simply increasing the number of sampled frames does not reliably improve accuracy. While proprietary models achieve the strongest overall performance, even leading systems struggle with global quality reasoning such as stability assessment and dominant factor attribution. These findings highlight that long-term perceptual reasoning remains an open challenge for current LVLMs, motivating future research toward more robust perceptual understanding over long-form videos.

% Please insert your acknowledgments here.

\section*{Acknowledgements}
This research is partially supported by the Ministry of Education, Singapore, under the funding of MOE-T2EP20123-0006. This work is also supported by gift funding from Amazon Prime Video for research on long-term video quality analysis. The authors would like to thank Alex Mackin and Benoit Vallade of Amazon Prime Video for their technical guidance and feedback on this research.

% ---- Bibliography ----
%
% BibTeX users should specify bibliography style 'splncs04'.
% References will then be sorted and formatted in the correct style.
%
\bibliographystyle{splncs04}
\bibliography{main}

\newpage

\begin{center}
    {\LARGE \textbf{Supplementary Material}}  
\end{center}

\appendix
\setcounter{figure}{0}
\renewcommand{\thefigure}{S\arabic{figure}}

\setcounter{table}{0}
\renewcommand{\thetable}{S\arabic{table}}

This supplementary document includes details on baseline implementations, experimental setup, and extended results for LongVQUBench.

\section{Baseline Implementation}

This section provides details of the evaluated baselines. In total, we evaluate \textbf{14 LVLMs}, comprising \textit{3 proprietary models}: GPT-5~\cite{openai2025gpt5}, Gemini~3~\cite{gemini3_google_2025}, and Qwen-VL-Max~\cite{bai2023qwen}; \textit{7 open-source models}: LLaVA-NeXT-Video~\cite{liu2024llavanext}, ShareGPT4Video~\cite{chen2024sharegpt4video}, Qwen3-VL~\cite{bai2025qwen3}, MovieChat~\cite{song2024moviechat}, LLaVA-Video~\cite{zhang2024llava}, VQA$^2$~\cite{jia2025vqa2}, and Long-RL~\cite{chen2025scaling}; and \textit{4 agentic LVLMs}: VideoAgent~\cite{VideoAgent}, VideoExplorer~\cite{yuan2025videoexplorer}, LongVT~\cite{yang2025longvt}, and DeepVideoDiscovery~\cite{zhang2025deep}. These models collectively cover proprietary, open-source, and agent-based paradigms, enabling a comprehensive analysis of long-video quality understanding capabilities of LVLMs. 

\subsection{Open-Source LVLMs}
We evaluate seven open-source long-video large vision-language models.

\begin{enumerate}

\item \textbf{LLaVA-NeXT-Video}~\cite{liu2024llavanext}\footnote{\url{https://github.com/LLaVA-VL/LLaVA-NeXT}} (HuggingFace: \texttt{lmms-lab/LLaVA-NeXT-Video-7B}) extends the LLaVA architecture from image understanding to video reasoning by processing multiple frames as visual tokens within a unified multimodal transformer. The model leverages a vision encoder and projector to map frame-level visual features into the language model space, enabling joint reasoning over visual and textual inputs. To support longer video inputs, the model adopts token-efficient frame representations and sequence length scaling techniques, allowing it to process longer frame sequences during inference. This design enables strong zero-shot video understanding capabilities without extensive video-specific training.

\item \textbf{ShareGPT4Video}~\cite{chen2024sharegpt4video}\footnote{\url{https://github.com/ShareGPT4Omni/ShareGPT4Video}} (HuggingFace: \texttt{Lin-Chen/sharegpt4video-8b}) focuses on improving video-language models through large-scale high-quality video caption supervision. The authors construct a dataset containing densely annotated video captions generated using GPT-4V, covering diverse video sources and durations. The model is trained using these dense temporal descriptions, enabling improved alignment between video frames and language. This approach significantly improves multi-frame reasoning and detailed video understanding compared to earlier video-language models.

\item \textbf{Qwen3-VL}~\cite{bai2025qwen3}\footnote{\url{https://github.com/QwenLM/Qwen3-VL}} (HuggingFace: \texttt{Qwen/Qwen3-VL-8B-Instruct}) is a multimodal extension of the Qwen large language model family designed for unified visual reasoning across images and videos. The model integrates a vision encoder with a large language backbone through cross-modal projection layers, enabling joint reasoning over visual tokens and text. It supports multiple visual inputs including images and video frames and demonstrates strong performance on multimodal reasoning, captioning, and visual question answering tasks.

\item \textbf{MovieChat}~\cite{song2024moviechat}\footnote{\url{https://github.com/rese1f/MovieChat}} (HuggingFace: \texttt{lmms-lab/MovieChat-ckpt}) is designed for long-video understanding and conversational reasoning over videos. The model introduces a hierarchical memory mechanism that compresses dense frame tokens into sparse memory representations, enabling efficient reasoning over long videos. This memory-based framework allows the model to maintain global context across extended temporal sequences while supporting interactive video question answering. The authors also introduce the MovieChat-1K benchmark for evaluation of long-video conversational understanding. MovieChat comprises approximately 8.2 billion parameters.

\item \textbf{LLaVA-Video}~\cite{zhang2024llava}\footnote{\url{https://github.com/LLaVA-VL/LLaVA-NeXT}} (HuggingFace: \texttt{lmms-lab/LLaVA-Video-7B-Qwen2}) extends the LLaVA framework to video inputs through video instruction tuning. The model processes sampled frames from videos and aligns them with textual instructions using multimodal instruction tuning. Synthetic video instruction datasets are used to improve the model’s ability to follow natural language queries related to video content. This approach allows the model to generalize from image instruction tuning to video reasoning tasks.

\item \textbf{VQA$^2$}~\cite{jia2025vqa2}\footnote{\url{https://github.com/Q-Future/Visual-Question-Answering-for-Video-Quality-Assessment}} (HuggingFace: \texttt{q-future/VQA-Assistant-llava-qwen-enhanced}) introduces the first large-scale instruction dataset and model suite that casts video quality assessment into a visual question answering paradigm, shifting from pure MOS prediction to joint scoring and understanding of video quality attributes. It was built on LLaVA-OneVision~\cite{li2024llava-onevision} and a SlowFast-R50~\cite{feichtenhofer2019slowfast} motion branch and has 8 billion parameters. It achieved state-of-the-art correlations on multiple UGC and streaming VQA benchmarks, while the VQA$^2$-Assistant interleaves visual and motion tokens to answer fine-grained quality understanding questions.

\item \textbf{Long-RL}~\cite{chen2025scaling}\footnote{\url{https://github.com/NVlabs/Long-RL}} (HuggingFace: \texttt{Efficient-Large-Model/LongVILA-R1-7B}) explores reinforcement learning strategies to improve long-video reasoning in LVLMs. The model uses reinforcement learning to optimize frame selection and reasoning trajectories during inference, allowing the system to focus on informative frames within long videos. This training paradigm improves temporal reasoning efficiency and scalability for long-duration video understanding tasks.

\end{enumerate}

\subsection{Closed-Source LVLMs}
We additionally evaluate three proprietary large vision-language models accessed through their respective APIs.

\begin{enumerate}

\item \textbf{GPT-5}~\cite{openai2025gpt5} is a proprietary multimodal model capable of reasoning across text, images, and videos. It integrates a large multimodal transformer architecture with extended context capabilities, enabling complex reasoning over long multimodal sequences. The model supports video analysis through sampled visual representations and demonstrates strong performance on multimodal reasoning and video understanding tasks.

\item \textbf{Gemini 3}~\cite{gemini3_google_2025} is a multimodal foundation model developed by Google that supports unified reasoning across text, images, audio, and video. The model is designed with long-context processing capabilities and advanced multimodal alignment mechanisms, enabling it to analyze long video sequences and perform complex reasoning tasks involving temporal dependencies.

\item \textbf{Qwen-VL-Max}~\cite{bai2023qwen} is a proprietary multimodal model from Alibaba’s Qwen family designed for high-performance visual reasoning tasks. The model integrates a large language backbone with a vision encoder and supports image and video understanding through a shared multimodal transformer. It demonstrates strong performance on visual question answering, captioning, and multimodal reasoning benchmarks.

\end{enumerate}

\subsection{Agentic LVLMs}
We further evaluate four agent-based video understanding systems that actively explore frames and perform multi-step reasoning.

\begin{enumerate}

\item \textbf{VideoAgent}~\cite{VideoAgent}\footnote{\url{https://github.com/wxh1996/VideoAgent}} is an agent-based framework that decomposes long-form video understanding into multiple reasoning steps. The system iteratively selects key frames or clips, analyzes visual evidence via tool calls, and updates intermediate reasoning states before generating the final answer. This agentic pipeline enables more targeted and efficient exploration of long videos compared to single-pass video models.

\item \textbf{VideoExplorer}~\cite{yuan2025videoexplorer}\footnote{\url{https://github.com/yhy-2000/VideoDeepResearch}} focuses on efficient frame exploration for long video reasoning. The framework dynamically selects informative frames based on the query and intermediate reasoning signals, reducing redundant visual processing. By combining retrieval-based frame selection with multimodal reasoning, the model improves efficiency for long video analysis.

\item \textbf{LongVT}~\cite{yang2025longvt}\footnote{\url{https://evolvinglmms-lab.github.io/LongVT/}} introduces a retrieval-based long-video reasoning framework that combines temporal grounding with multimodal reasoning modules. The model identifies relevant segments from long videos via global-to-local temporal selection and aggregates the retrieved visual evidence before generating answers. This design improves scalability when reasoning over very long videos.

\item \textbf{DeepVideoDiscovery}~\cite{zhang2025deep}\footnote{\url{https://github.com/microsoft/DeepVideoDiscovery}} is an agentic system designed for adaptive exploration of long videos. The framework iteratively discovers informative frames using a reasoning-guided search strategy and integrates them into a multimodal reasoning pipeline. This iterative discovery process enables improved temporal reasoning and strong performance on long-video understanding benchmarks.

\end{enumerate}

\begin{figure}[t]
\centering
\includegraphics[width=\linewidth, trim={1cm 8cm 5cm 3cm}, clip]{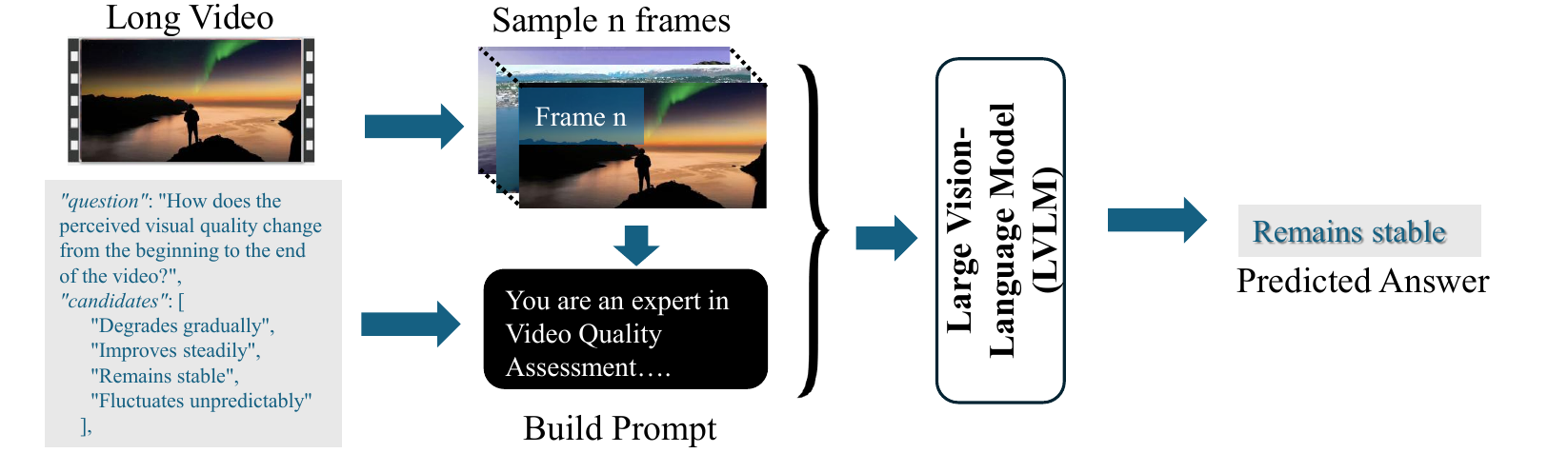}
\caption{Inference Pipeline}
\label{fig:inference_pipeline}
\end{figure}

\section{Experimental Setup}
This section describes the evaluation protocol used in our experiments,
including the inference pipeline for processing long-form videos, extended results and the
prompt formats used for multiple-choice and open-ended questions.

\subsection{Inference Pipeline}
For each long-form video, we first perform uniform frame sampling at 1 FPS
over the entire duration, obtaining a temporally ordered sequence of frames
that approximates the full viewing experience. For every question associated
with that video (either multiple-choice or open-ended), we then construct a
question-specific prompt by combining a shared video-context header (including
the video duration, sampling rate, and number of sampled frames) with the
question text and, for multiple-choice items, the corresponding answer
candidates. Finally, as illustrated in Figure~\ref{fig:inference_pipeline}, the sampled frames and the constructed prompt are jointly
fed into the LVLM in a single forward pass, which returns either a discrete
option label for multiple-choice questions or a concise textual response for
open-ended questions.

\subsection{Prompt Format}

For each video, we first construct a shared video-context header based on
its duration and the number of sampled frames:
{\color{blue!70!black}
\begin{verbatim}
You are an expert in Video Quality Understanding.

The video is <VIDEO_DURATION> seconds long. 
Frames were sampled at  1 FPS (frames per second). 
<NUM_FRAMES> frames uniformly selected across the video duration. 
Frames are in chronological order from start to end.  
\end{verbatim}
}

\paragraph{Multiple-choice questions.}
Given a question and its candidate answers, we construct the MCQ
prompt as:
{\color{blue!70!black}
\begin{verbatim}
You are an expert in Video Quality Understanding.

The video is <VIDEO_DURATION> seconds long. 
Frames were sampled at  1 FPS (frames per second). 
<NUM_FRAMES> frames uniformly selected across the video duration. 
Frames are in chronological order from start to end.  

Question:
<QUESTION_TEXT>

Choices:
A. <CANDIDATE_1>
B. <CANDIDATE_2>
...

Select the correct answer.

IMPORTANT:
Return ONLY one letter from: A, B, C, D.
Do NOT give extra text description in answer.
\end{verbatim}
}
\paragraph{Open-ended questions.}
For open-ended questions, we use:
{\color{blue!70!black}
\begin{verbatim}
You are an expert in Video Quality Understanding.

The video is <VIDEO_DURATION> seconds long. 
Frames were sampled at  1 FPS (frames per second). 
<NUM_FRAMES> frames uniformly selected across the video duration. 
Frames are in chronological order from start to end.  

Question:
<QUESTION_TEXT>

Give a descriptive answer (maximum 80 words).
\end{verbatim}
}

\subsection{Extended Results}

\noindent\textbf{Synthetic vs.\ Authentic Distortions.}
\label{r1:q2}
Synthetic distortions are applied to 888 high-quality videos (400 LQU, 400 CQR, 88 GQU) for reliable QA 
construction, while the 218 medium- and 94 low-quality GQU videos contain authentic 
distortions (Main Paper - Sec.~3.2). As shown in Table~\ref{tab:syn_versus_auth}, all LVLMs 
consistently perform better on synthetic than authentic distortions, confirming that 
real-world degradations pose greater challenges for current models.

\begin{table}[h]
\centering
\setlength{\tabcolsep}{1.5pt}
\renewcommand{\arraystretch}{1.1}
\caption{Performance (\%) on synthetic (syn) vs.\ authentic (ath) distortions across LQU, CQR, and GQU evaluation levels. GQU-H/M/L denotes high/medium/low quality GQU videos.}
\resizebox{0.85\textwidth}{!}{
\begin{tabular}{ccccccccc}
\toprule
\multirow{2}{*}{\textbf{Model}} & \multicolumn{4}{c}{\textbf{Synthetic}} & \multicolumn{3}{c}{\textbf{Authentic}} \\
\cmidrule(lr){2-5} \cmidrule(lr){6-8}
 & \textbf{LQU} & \textbf{CQR} & \textbf{GQU-H} & \textbf{Overall$_\text{syn}$} 
               & \textbf{GQU-M} & \textbf{GQU-L} & \textbf{Overall$_\text{ath}$} \\
\midrule
GPT-5              & 65.4 & 82.2 & 66.1 & 71.76 & 55.4 & 61.2 & 57.15 \\
LongRL             & 61.9 & 67.1 & 56.2 & 62.30 & 45.4 & 50.9 & 47.06 \\
DeepVideoDiscovery & 69.2 & 72.1 & 60.8 & 68.04 & 53.8 & 55.8 & 54.40 \\
\bottomrule
\end{tabular}
}
\label{tab:syn_versus_auth}
\end{table}

\noindent\textbf{Sampling Rate Ablation.}
\label{r1:q3}
As shown in Table~\ref{tab:sampling_ablation}, increasing the sampling rate does not 
consistently improve performance across any of the three hierarchical evaluation levels. GPT-5 and LongRL 
remain stable across all FPS settings, as both models uniformly subsample 
$\#\text{max\_frames}$ after initial video loading, making their outputs largely 
independent of FPS. DeepVideoDiscovery exhibits similar insensitivity to FPS changes. 
These results justify our use of the 1~FPS setting, which is consistent with prior 
video quality understanding benchmarks~\cite{zhang2025qbench}.

\begin{table}[h]
\centering
\setlength{\tabcolsep}{5pt}
\renewcommand{\arraystretch}{1.1}
\caption{Performance (\%) under varying sampling rates (1, 8, 16 FPS) across LQU, CQR, and GQU evaluation levels.}
\resizebox{0.8\linewidth}{!}{%
\begin{tabular}{c ccc ccc ccc}
\toprule
\multirow{2}{*}{\textbf{Model}} & \multicolumn{3}{c}{\textbf{LQU}}
& \multicolumn{3}{c}{\textbf{CQR}}
& \multicolumn{3}{c}{\textbf{GQU}} \\
\cmidrule(lr){2-4} \cmidrule(lr){5-7} \cmidrule(lr){8-10}
 & \textbf{1} & \textbf{8} & \textbf{16}
              & \textbf{1} & \textbf{8} & \textbf{16}
              & \textbf{1} & \textbf{8} & \textbf{16} \\
\midrule
GPT-5
  & 65.4 & 65.2 & 64.8
  & 82.2 & 81.9 & 82.1
  & 60.9 & 60.4 & 60.7 \\
LongRL
  & 61.9 & 61.6 & 61.0
  & 67.1 & 66.8 & 67.3
  & 50.8 & 49.9 & 50.5 \\
DeepVideoDiscovery
  & 69.2 & 69.7 & 69.5
  & 72.1 & 72.1 & 72.3
  & 56.8 & 56.4 & 56.8 \\
\bottomrule
\end{tabular}
}
\label{tab:sampling_ablation}
\end{table}

\section{Dataset Construction}
This section presents additional details of the dataset
construction process, including video-question pair distribution, duration distribution, controlled distortion setup, and the verification pipeline for question–answer pairs.

%%%%%%%%%%%%%%%%
\subsection{Extended Dataset Statistics}

\noindent
\textbf{Video-Question Distribution.}
LongVQUBench adopts a hierarchical structure to evaluate long video quality 
understanding at increasing levels of complexity, as summarized in 
Table~\ref{tab:distribution}. The first level, Local Quality Understanding (LQU), 
focuses on fine-grained distortion perception within individual frames or short segments. 
The second level, Cross-event Quality Reasoning (CQR), requires models to reason 
across multiple temporal segments, assessing quality comparison, cumulative effects, and 
temporal relationships. The third level, Global Quality Understanding (GQU), 
evaluates holistic video quality, including stability, dominant distortion factors, and 
overall quality assessment. Each level comprises 400 videos and 500 questions (100 per perceptual
dimension), yielding 1200 videos and 1500 questions.

\begin{table}[h]
\centering
\caption{Video-Question distribution across LQU, CQR, and GQU evaluation levels.}
\setlength{\tabcolsep}{4pt}
\renewcommand{\arraystretch}{1.1}
\resizebox{0.6\linewidth}{!}{%
\begin{tabular}{clcc}
\toprule
\textbf{Level} & \textbf{Dimension} & \textbf{\#Videos} & \textbf{\#Q} \\
\midrule
\multirow{5}{*}{LQU}
  & Detection            & \multirow{5}{*}{400} & 100 \\
  & Localization         &                      & 100 \\
  & Classification       &                      & 100 \\
  & Severity \& Comfort  &                      & 100 \\
  & Open-Ended           &                      & 100 \\
\midrule
\multirow{5}{*}{CQR}
  & Comparison           & \multirow{5}{*}{400} & 100 \\
  & Cumulative Effect    &                      & 100 \\
  & Integration          &                      & 100 \\
  & Temporal Relation    &                      & 100 \\
  & Open-Ended           &                      & 100 \\
\midrule
\multirow{5}{*}{GQU}
  & Stability \& Consistency & \multirow{5}{*}{400} & 100 \\
  & Dominant Factors         &                      & 100 \\
  & Temporal Trend           &                      & 100 \\
  & Overall Quality          &                      & 100 \\
  & Open-Ended               &                      & 100 \\
\midrule
\textbf{Total} & 15 subcategories & \textbf{1200} & \textbf{1500} \\
\bottomrule
\end{tabular}
}
\label{tab:distribution}
\end{table}

\noindent
\textbf{Video Duration Distribution.}
Table~\ref{tab:level_duration_distribution} presents the video duration distribution across 
the three evaluation levels. LongVQUBench provides broad temporal coverage, 
with videos ranging from under $1.5+$ minutes to $\sim$2 hours across all levels. 
LQU and GQU videos are well distributed across the 3--30 minute range, 
while CQR videos have stronger representation in longer durations (8--30 minutes), 
naturally aligning with the presence of multiple distinct degradations. 
The three levels (LQU, CQR, and GQU) cover all duration ranges, 
ensuring a comprehensive evaluation of long video quality understanding.

\begin{table}[h]
\centering
\setlength{\tabcolsep}{5pt}
\renewcommand{\arraystretch}{1.1}
\caption{Video duration distribution across LQU, CQR, and GQU evaluation levels.}
\resizebox{0.7\linewidth}{!}{%
\begin{tabular}{ccccccc}
\toprule
\textbf{Duration (mins)} & \textbf{0--3} & \textbf{3--5} & \textbf{5--8} & 
\textbf{8--15} & \textbf{15--30} & \textbf{30--120} \\
\midrule
LQU & 0   & 81  & 103 & 84  & 107 & 25 \\
CQR & 22  & 19  & 61  & 149 & 135 & 14 \\
GQU & 3   & 73  & 136 & 113 & 57  & 18 \\
\midrule
Total & 25 & 173 & 300 & 
346 & 299 & 57 \\
\bottomrule
\end{tabular} }
\label{tab:level_duration_distribution}
\end{table}
%%%%%%%%%%%%

\subsection{Controlled Distortion Configuration}

To systematically evaluate LVLM performance under varying video quality, we applied a set of spatial and temporal distortions to the videos in LongVQUBench. Spatial distortions affect individual frames, while temporal distortions affect frame sequences (clips). Each distortion is applied at multiple intensity levels to simulate varying severity or visibility.

\begin{table}
\centering
\caption{Distortion types applied to LongVQUBench. Each level lists the distortion intensity and the number of affected videos in separate subcolumns.}
\resizebox{0.95\linewidth}{!}{
\begin{tabular}{c|cc|cc|cc}
\toprule
\textbf{Distortion} & \multicolumn{2}{c|}{\textbf{Level 1}} & \multicolumn{2}{c|}{\textbf{Level 2}} & \multicolumn{2}{c}{\textbf{Level 3}} \\
 & \textbf{Intensity} & \textbf{\#Videos} & \textbf{Intensity} & \textbf{\#Videos} & \textbf{Intensity} & \textbf{\#Videos} \\
\midrule
\rowcolor{gray!20}
\multicolumn{7}{c}{\textit{Spatial Distortions}} \\
\midrule

Brightness Increase & 30 & 151 & 80 & 89 & 150 & 164 \\
Contrast Reduction & 0.8 & 107 & 0.4 & 138 & 0.2 & 172 \\
Defocus Blur & 10 & 92 & 25 & 158 & 50 & 24 \\
Gaussian Blur & 7 & 235 & 21 & 68 & 45 & 65 \\
Gaussian Noise & 15 & 64 & 30 & 82 & 80 & 24 \\
Hue Shift & 15 & 175 & 60 & 56 & 130 & 87 \\
JPEG Compression & 30 & 117 & 10 & 22 & 3 & 111 \\
Motion Blur & 10 & 74 & 25 & 118 & 50 & 129 \\
Pixelation & 10 & 95 & 70 & 25 & 130 & 25 \\
Poisson Noise & \multicolumn{6}{c}{No intensity level, \#Videos=395}  \\
Salt \& Pepper Noise & 0.03 & 70 & 0.10 & 35 & 0.30 & 139 \\
Saturation Shift & 0.8 & 153 & 2.0 & 109 & 4.0 & 147 \\
Sharpening Artifacts & 2.0 & 86 & 6.0 & 56 & 12.0 & 82 \\
Speckle Noise & 0.1 & 191 & 0.4 & 123 & 0.8 & 175 \\

\midrule
\rowcolor{gray!20}
\multicolumn{7}{c}{\textit{Temporal Distortions}} \\
\midrule

Flicker & 0.2 & 147 & 0.7 & 158 & 1.2 & 112 \\
Frame Drop & 0.1 & 137 & 0.4 & 149 & 0.7 & 150 \\
Jitter & 5 & 168 & 15 & 175 & 30 & 148 \\
Stutter & 5 & 173 & 15 & 154 & 25 & 143 \\

\bottomrule
\end{tabular}
}
\label{tab:distortions}
\end{table} % use input not \include

\noindent
\textbf{Spatial Distortions:} These distortions affect individual frames and simulate common video artifacts or manipulations. The distortion levels denote different parameter settings or variants and do not necessarily correspond to monotonically increasing severity.

\begin{figure}
\centering
\includegraphics[width=0.95\linewidth, trim=0cm 0.25cm 0cm 0cm, clip]
{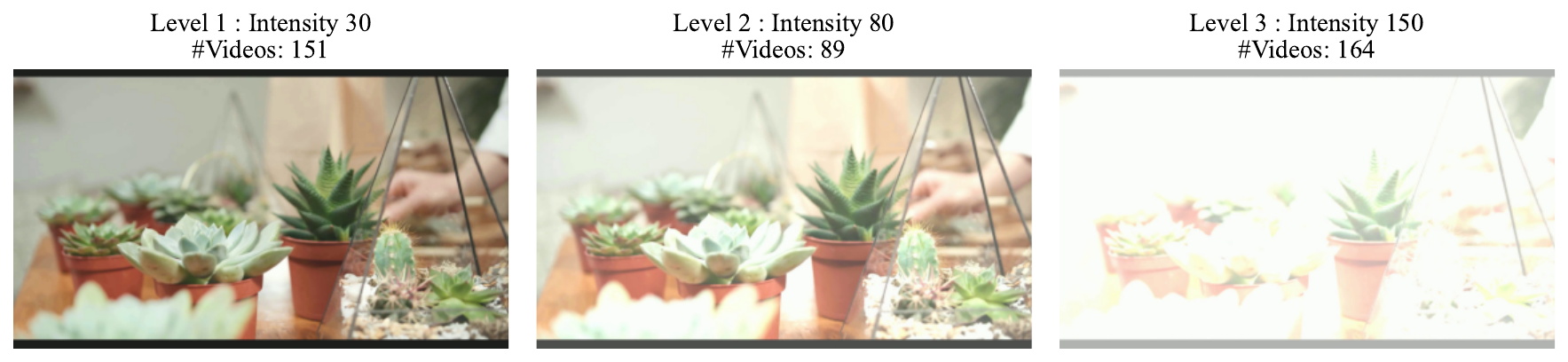}
\caption{Illustration of brightness increase across different intensity levels.}
\label{fig:brightness}
\end{figure}

\begin{figure}
\includegraphics[width=0.95\linewidth, trim=0cm 0.25cm 0cm 0cm, clip]
{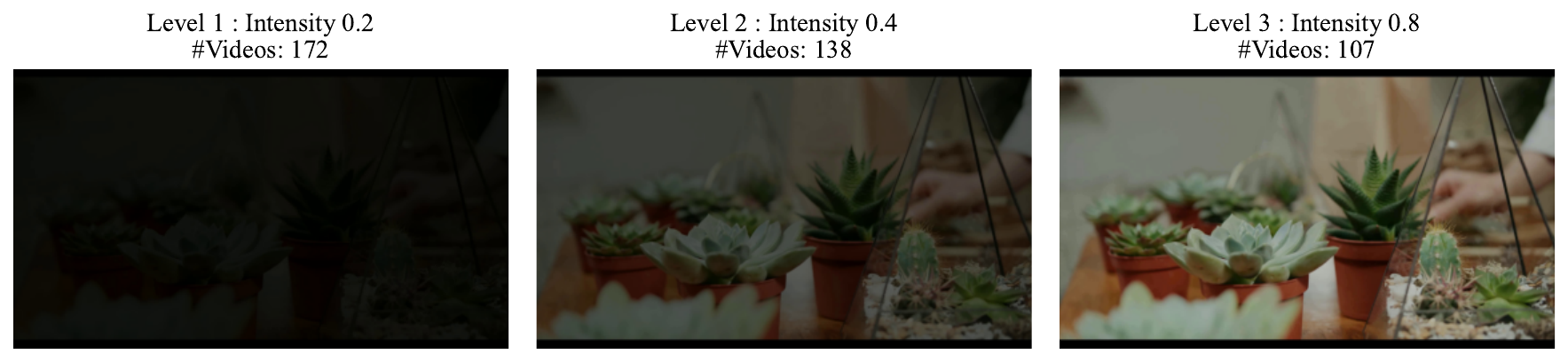}
\caption{Illustration of contrast reduction across different intensity levels.}
\label{fig:contrast}
\end{figure}

\begin{figure}
\centering
\includegraphics[width=0.95\linewidth, trim=0cm 0.25cm 0cm 0cm, clip]
{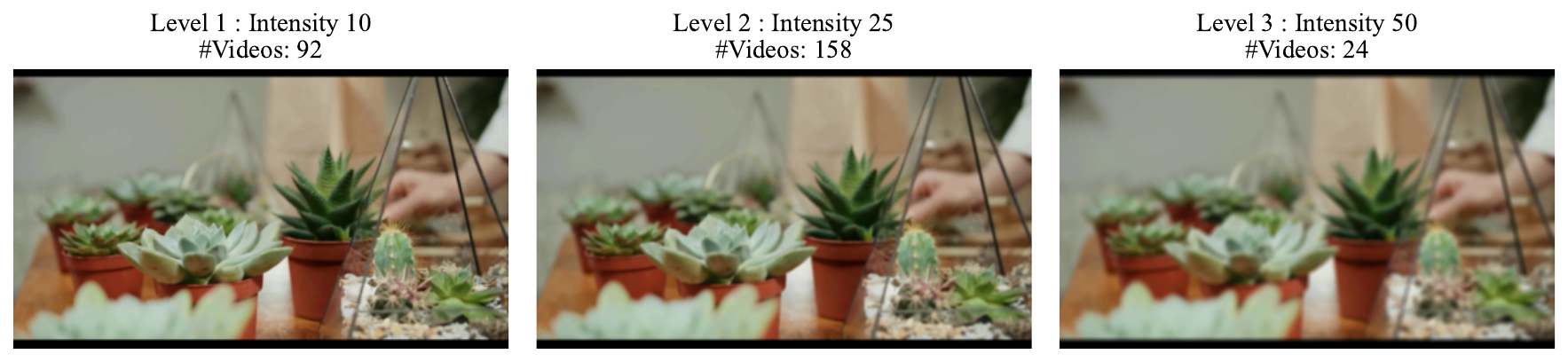}
\caption{Illustration of defocus blur across different intensity levels.}
\label{fig:defocus_blur}
\end{figure}

\begin{figure} 
\centering
\includegraphics[width=0.95\linewidth, trim=0cm 0.25cm 0cm 0cm, clip]
{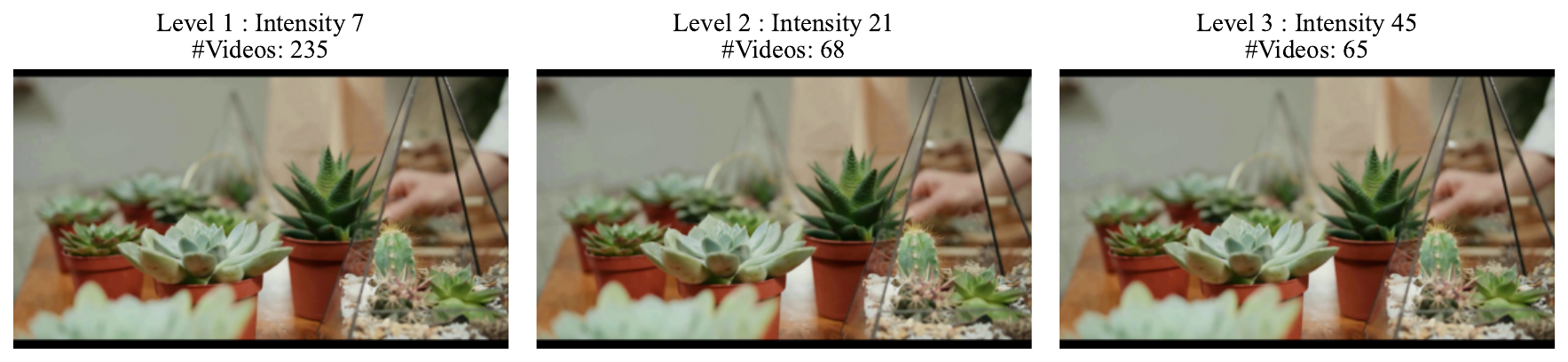}
\caption{Illustration of Gaussian blur across different intensity levels.}
\label{fig:gaussian_blur}
\end{figure}

\begin{figure} 
\centering
\includegraphics[width=0.95\linewidth, trim=0cm 0.25cm 0cm 0cm, clip]
{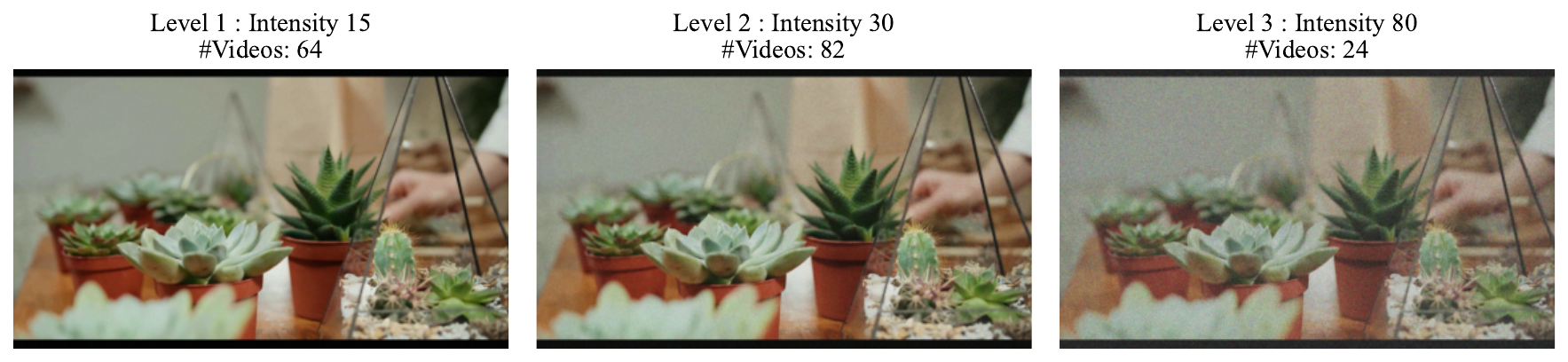}
\caption{Illustration of Gaussian noise across different intensity levels.}
\label{fig:gaussian_noise}
\end{figure}

\begin{figure} 
\centering
\includegraphics[width=0.95\linewidth, trim=0cm 0.25cm 0cm 0cm, clip]
{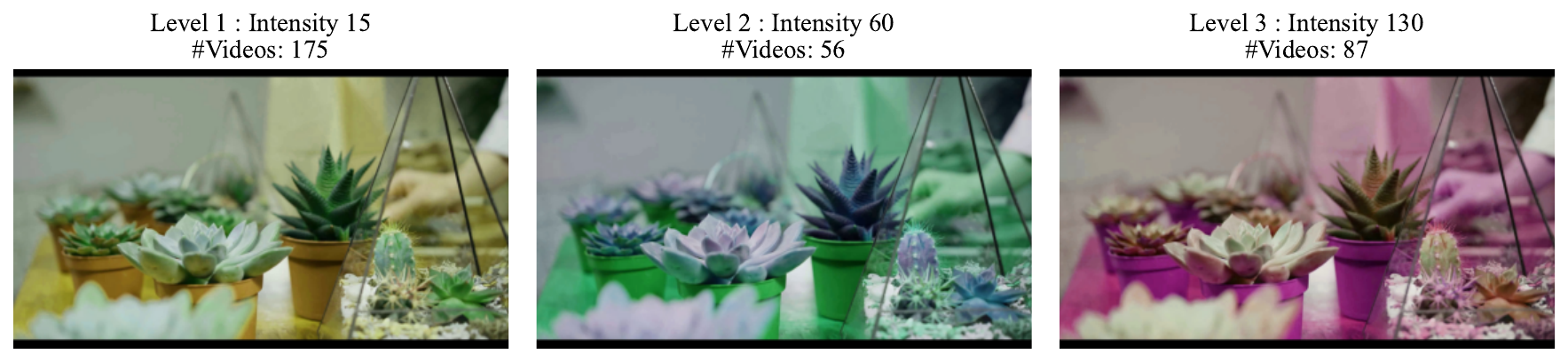}
\caption{Illustration of hue shift across different intensity levels.}
\label{fig:hue}
\end{figure}

\begin{figure} 
\centering
\includegraphics[width=0.95\linewidth, trim=0cm 0.25cm 0cm 0cm, clip]
{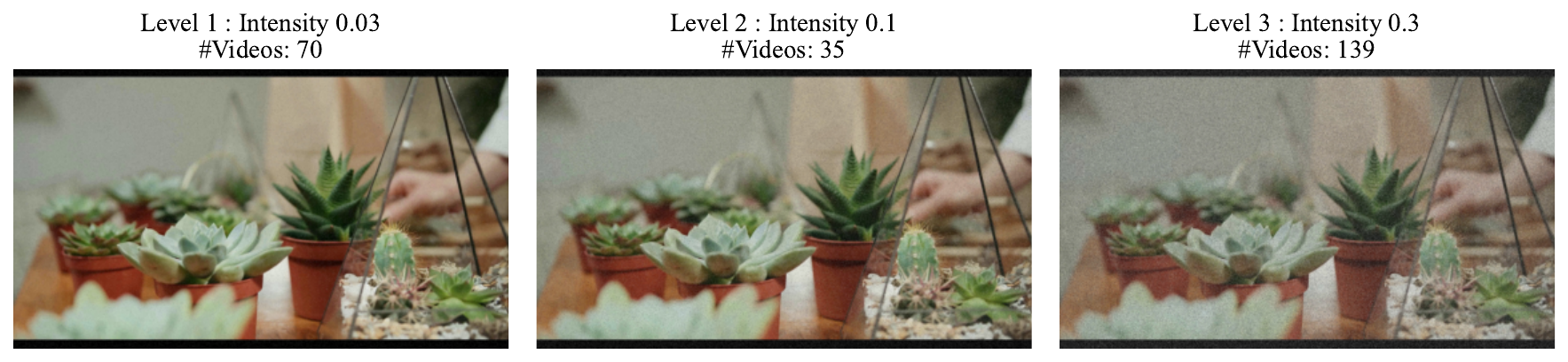}
\caption{Illustration of salt-and-pepper noise across different intensity levels.}
\label{fig:salt_pepper_noise}
\end{figure}

\begin{figure} 
\centering
\includegraphics[width=0.95\linewidth, trim=0cm 0.25cm 0cm 0cm, clip]
{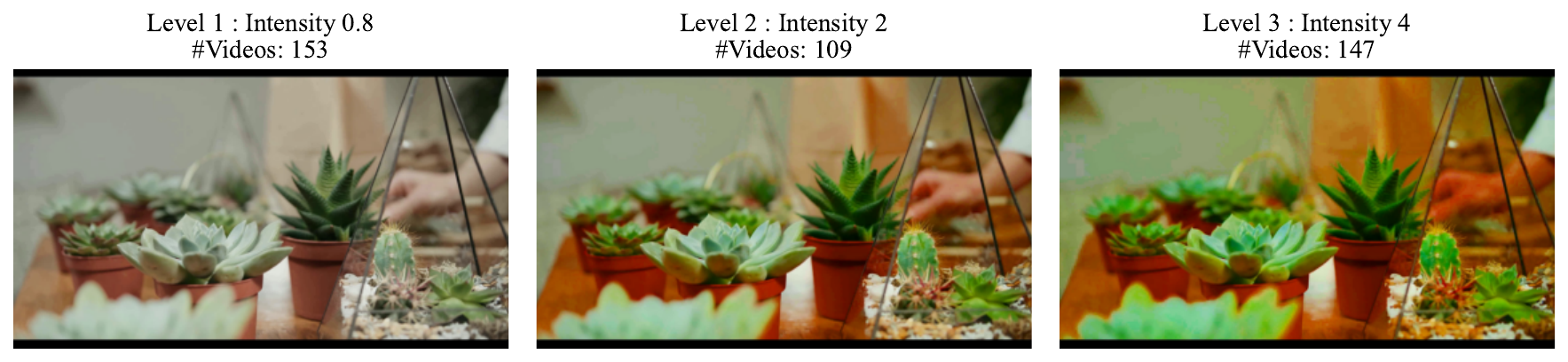}
\caption{Illustration of saturation shift across different intensity levels.}
\label{fig:saturation}
\end{figure}

\begin{figure} 
\centering
\includegraphics[width=0.95\linewidth, trim=0cm 0.25cm 0cm 0cm, clip]
{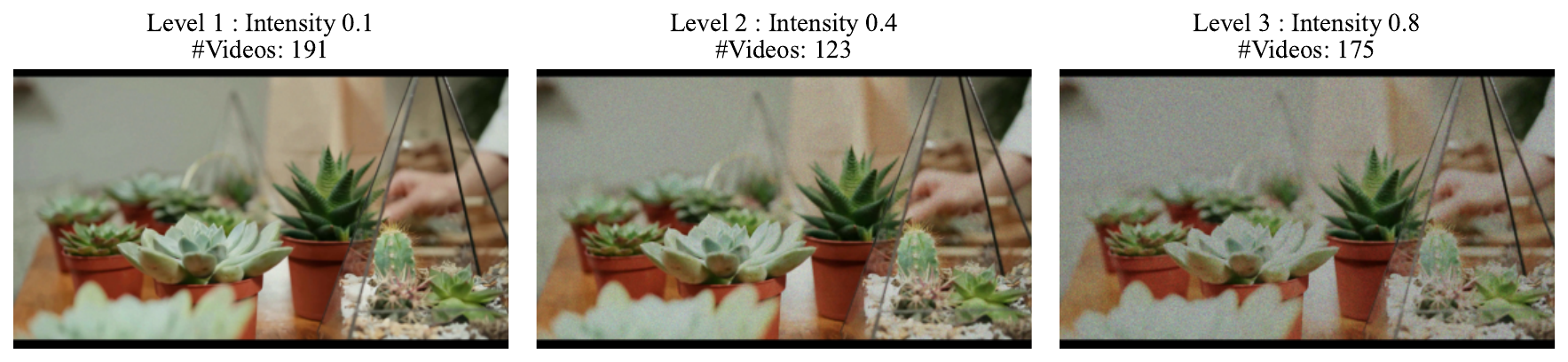}
\caption{Illustration of speckle noise  across different intensity levels.}
\label{fig:speckle_noise}
\end{figure}

\begin{figure} 
\centering
\includegraphics[width=0.35\linewidth, trim=0cm 0.25cm 0cm 0.8cm, clip]
{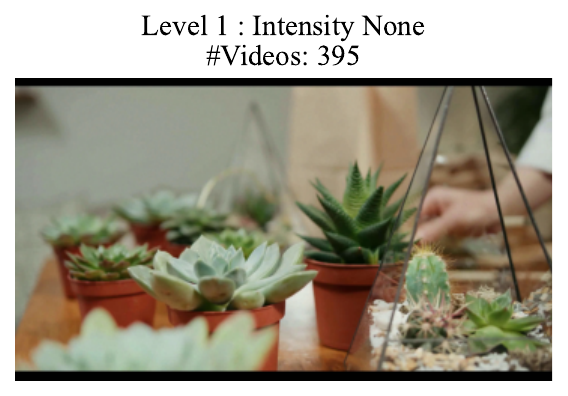}
\caption{Illustration of Poisson noise distortion. No controllable levels.}
\label{fig:poisson_noise}
\end{figure}

\begin{figure} 
\centering
\includegraphics[width=0.95\linewidth, trim=0cm 0.25cm 0cm 0cm, clip]
{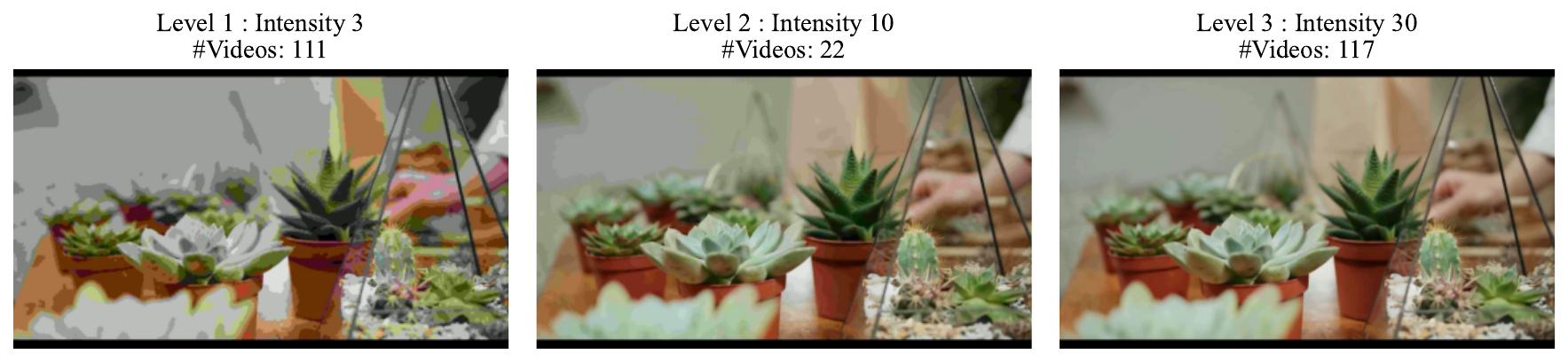}
\caption{Illustration of JPEG compression across different intensity levels.}
\label{fig:jpeg_compression}
\end{figure}

\begin{enumerate}

    \item \textbf{Brightness Increase} -- Increases frame brightness (see Figure~\ref{fig:brightness}). Intensity 30 is mildly brighter, 80 is blown out, and 150 approaches almost white frames.
    
    \item \textbf{Contrast Reduction} -- Reduces frame contrast (see Figure~\ref{fig:contrast}). Intensity 0.8 is slightly dull, 0.4 is washed out, and 0.2 nearly flattens contrast (almost black).

    \item \textbf{Defocus Blur} -- Simulates optical defocus (see Figure~\ref{fig:defocus_blur}). Intensity 10 results in slight blur, 25 produces smeared frames, and 50 creates foggy frames.

    \item \textbf{Gaussian Blur} -- Smooths the image (see Figure~\ref{fig:gaussian_blur}). Intensity 7 is very mild softening, 21 is moderate, and 45 is strong blur.

    \item \textbf{Gaussian Noise} -- Adds random pixel noise (see Figure~\ref{fig:gaussian_noise}). Intensity 15 is light noise, intensity 30 is heavier grain, and intensity 80 resembles a sandstorm-like appearance.

    \item \textbf{Hue Shift} -- Rotates colors in the hue space (see Figure~\ref{fig:hue}). Intensity 15 gives a small color shift, intensity 60 results in a strong tint, and intensity 130 produces unnatural, “alien” colors. Note: This is rotational rather than intensity-based.

    \item \textbf{JPEG Compression} -- Introduces block artifacts and information loss (see Figure~\ref{fig:jpeg_compression}). Intensity 30 corresponds to mild blocking,  10 to heavy compression with noticeable quality loss, and  3 results in broken fine details.

    \item \textbf{Motion Blur} -- Introduces streaking due to simulated motion. Intensity 10 is slight trails, 25 shows long streaks, and 50 produces full smear.

    \item \textbf{Pixelation} -- Reduces spatial resolution by blockification. Intensity 10 gives small blocks, 70 is clearly visible blocks, and 130 resembles large block appearance.

    \item \textbf{Poisson Noise} -- Simulates photon shot noise (see Figure~\ref{fig:poisson_noise}). This distortion does not have controllable level intensity and is inherently strong.

    \item \textbf{Salt \& Pepper Noise} -- Random black and white pixels (see Figure~\ref{fig:salt_pepper_noise}). Intensity 0.03 produces few sparkles, 0.10 generates noticeable impulses, and 0.30 creates broken frames.

    \item \textbf{Saturation Shift} -- Modifies color vividness (see Figure~\ref{fig:saturation}). Intensity 0.8 slightly reduces saturation, 2.0 produces neon-like colors, and 4.0 is unrealistic saturation in frame. 

    \item \textbf{Sharpening Artifacts} -- Adds halo and ringing artifacts. Intensity 2 creates slight halos, 6 produces noticeable ringing, and 12 generates harsh outlines.

    \item \textbf{Speckle Noise} -- Multiplicative noise creating speckled patterns (see Figure~\ref{fig:speckle_noise}). Intensity 0.1 gives light specks, 0.4 produces stronger “snowy” patterns, and 0.8 is massive disturbance.

\end{enumerate}

\noindent
\textbf{Temporal Distortions:} These distortions affect frame sequences and simulate playback issues or unstable captures.

\begin{enumerate}

    \item \textbf{Flicker} -- 
    Video flicker occurs when the camera frame rate is not synchronized with the lighting frequency (e.g., 50/60 Hz), producing periodic bright and dark bands. This artifact commonly arises under artificial lighting or when recording digital screens. Intensity 0.2 produces mild flashes, 0.7 strong flicker, and 1.2 creates a strobe-like effect.

    % \item \textbf{Frame Drop} -- Randomly removes frames. Level 0.1 is barely noticeable, 0.4 produces choppiness, and 0.7 creates a slideshow-like effect.

    % \item \textbf{Jitter} -- Adds small frame displacements or shaking. Level 5 gives slight shake, 15 is more unstable, and 30 produces strong frame instability.

    % \item \textbf{Stutter} -- Repeats or freezes frames. Level 5 introduces small freezes, 15 creates noticeable pauses, and 25 results in long freezes.

    \item \textbf{Frame Drop} -- Randomly removes frames from the video stream. Intensity 0.1 is barely noticeable, 0.4 produces visible choppiness, and 0.7 creates a slideshow-like effect. This distortion disrupts temporal continuity and can hinder motion perception.

\item \textbf{Jitter} -- Adds small random frame displacements or shaking. Intensity 5 introduces slight camera shake, 15 results in noticeable instability, and 30 produces strong frame jitter. This distortion simulates unstable capture conditions such as handheld recording.

\item \textbf{Stutter} -- Repeats or freezes frames intermittently. Intensity 5 introduces short freezes, 15 creates noticeable pauses, and 25 results in long freezing artifacts. This distortion disrupts smooth motion playback and creates temporal discontinuities.

\end{enumerate}

\subsection{Question--Answer Pair}
This section describes the verification workflow used to construct the
question-answer (QA) pairs in the dataset, including interface
and verification procedure used to ensure correctness and consistency of the questions.

\begin{figure}
\centering
\includegraphics[width=\linewidth, trim={0.3cm 0.5cm 0.5cm 2.8cm}, clip]{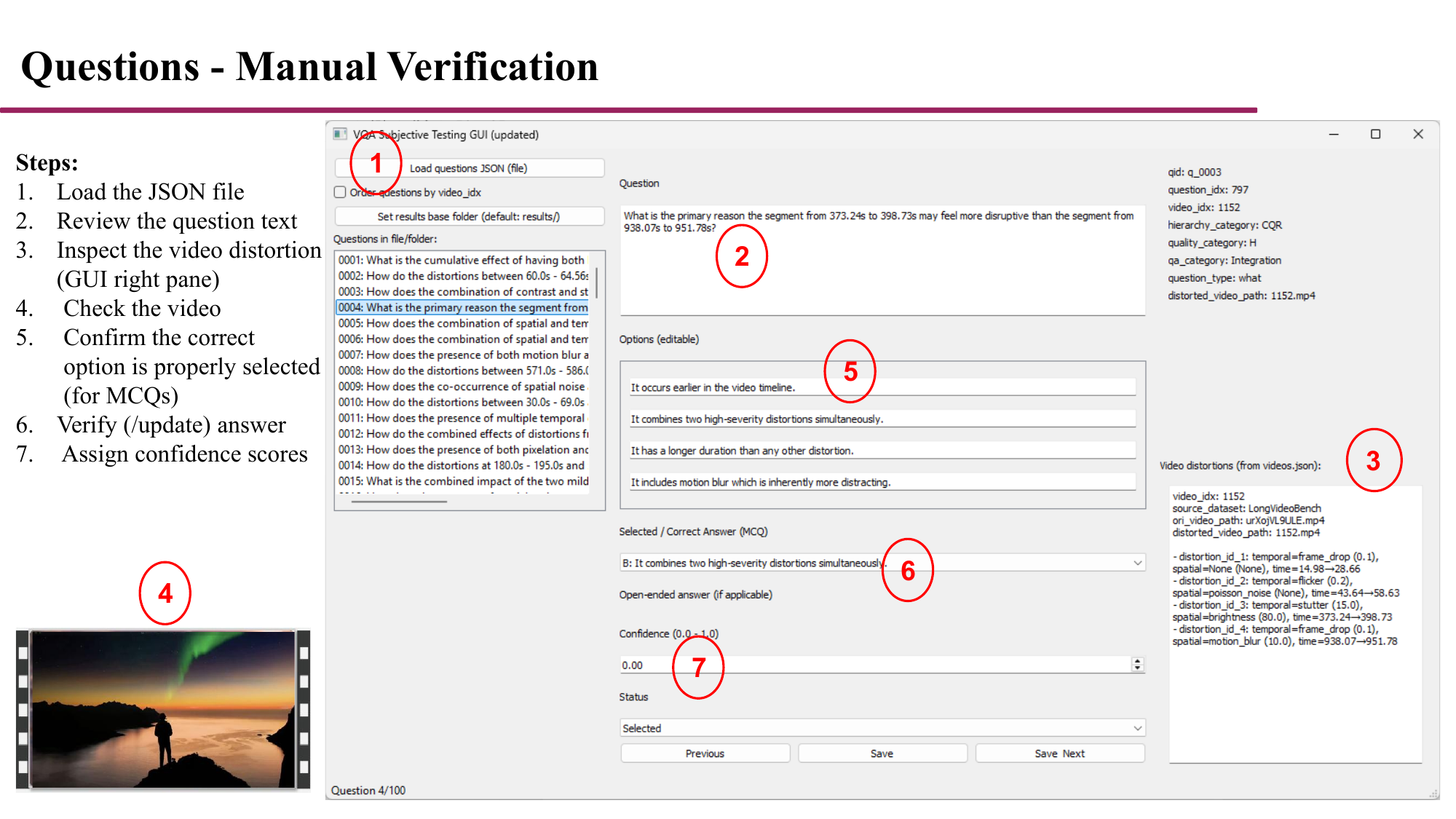}
\caption{Question verification steps within the GUI, where annotators review generated questions and validate their correctness before inclusion in the dataset.}
\label{fig:gui}
\end{figure}

\noindent
\textbf{Question Verification GUI.}
The verification interface shown in Figure~\ref{fig:gui} was developed to facilitate efficient verification and refinement of QA pairs. The GUI allows annotators to load a video together with its corresponding JSON file containing pre-generated questions and answers. These questions automatically populate the relevant fields in the interface, enabling annotators to quickly review them in context with the video. All question and answer fields are editable, allowing annotators to modify wording, adjust answer options, or correct labels when necessary. This design enables rapid iteration over the QA set while ensuring that questions remain aligned with the visual content of the video. 
%To further improve the quality and consistency of the dataset, we performed question verification in four iterative rounds. In each round, annotators reviewed and refined the existing questions, resolving ambiguities and correcting errors. Figure~\ref{fig:question_update} illustrates how the set of questions evolved and stabilized across these verification iterations.

\begin{figure}
\centering
\includegraphics[width=\linewidth, trim={2cm 10cm 3cm 1.8cm}, clip]{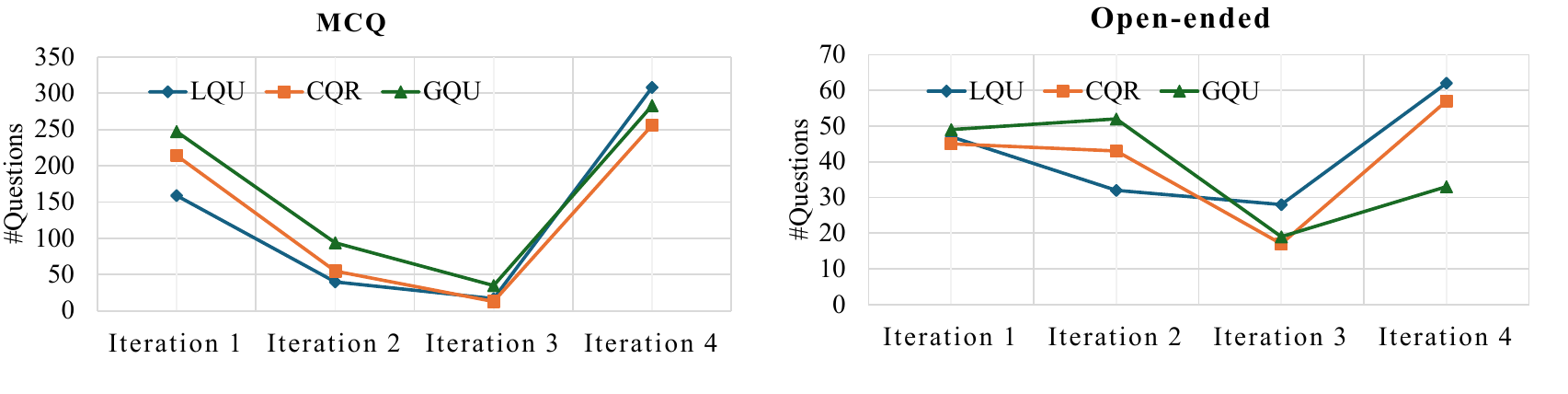}
\caption{Number of questions updated in each iteration.}
\label{fig:question_update}
\end{figure}

\noindent
\textbf{Iterative Question Verification.}
QA verification was conducted in multiple rounds using the custom GUI described above. During iterations~1--3, non-expert verifiers loaded the corresponding JSON verification file and reviewed each candidate QA pair while watching the associated video. The GUI automatically populated the question and answer fields, which verifiers could edit to correct wording, adjust answer options, or refine the correct label.  In iteration~4, experts in multimedia quality assessment systematically revisited the questions to improve clarity, resolve ambiguities, and ensure that each question accurately reflected the visible distortion in the referenced video segment. Ambiguous or poorly specified questions were revised, while additional cues, such as more precise temporal references, were added when necessary. Figure~\ref{fig:question_update} illustrates how the QA set evolved across the four verification rounds, showing the progressive refinement of the questions. Through this iterative process, we obtained the final set of 1,500 QA pairs with improved clarity, temporal grounding, and consistency across all three hierarchical evaluation levels - LQU, CQR and GQU.

\section{Evaluation of Open-Ended Responses}
% We employed GPT-5~\cite{openai2025gpt5}-based scoring to evaluate open-ended questions inspired by MLVU~\cite{zhao2025mlvu}. We formulated the prompt shown in the following section to assign a score to each response and . 

We evaluate open-ended question answering using two complementary metrics. First, we employ a GPT-5-based prompt evaluation framework, inspired by MLVU~\cite{zhao2025mlvu}, to assess the \emph{relevance} and \emph{completeness} of each generated response. Second, we report BERTScore-F1~\cite{zhang2019bertscore} as a semantic similarity metric between the generated and reference answers.

\subsection{BERTScore Evaluation}

To complement the LLM-based relevance and completeness evaluation, we additionally report BERTScore-F1 as an automatic semantic similarity metric for open-ended question answering. Unlike traditional n-gram-based metrics such as BLEU~\cite{papineni2002bleu}, BERTScore~\cite{zhang2019bertscore} measures semantic similarity by comparing contextualized token embeddings generated by a pretrained language model. Given a predicted answer and its corresponding reference answer, BERTScore computes pairwise cosine similarities between token embeddings, from which precision and recall are estimated through greedy token matching. The final BERTScore-F1 is computed as the harmonic mean of precision and recall, providing a robust measure of semantic overlap even when equivalent information is expressed using different wording. Table~\ref{tab:bertscore_f1} reports the BERTScore-F1 results on the {\tt Test} set. The results exhibit trends consistent with the GPT-5-based prompt evaluation.

\begin{table}[t]
\centering
\setlength{\tabcolsep}{7pt}
\caption{Leaderboard of BERTScore-F1 on open-ended questions in the {\tt Test} set.}
\resizebox{0.7\linewidth}{!}{%
\begin{tabular}{ccccc}
\toprule
\textbf{Model} & \textbf{LQU} & \textbf{CQR} & \textbf{GQU} & \textbf{Overall} \\
\midrule
GPT-5 & 0.5877 & 0.6118 & 0.5829 & 0.5941 \\
Long-RL & 0.4942 & 0.5227 & 0.4982 & 0.5050 \\
DeepVideoDiscovery & 0.5168 & 0.5554 & 0.5347 & 0.5356 \\
\bottomrule
\end{tabular}
}
\label{tab:bertscore_f1}
\end{table}

\subsection{GPT-based Prompt for Open-Ended Evaluation}

Following prior work on automated evaluation for video understanding benchmarks such as MLVU~\cite{zhao2025mlvu}, we employ a GPT-5~\cite{openai2025gpt5}-based evaluator to score responses to open-ended questions. Unlike multiple-choice questions, open-ended responses may vary in phrasing while still conveying correct information. Therefore, direct string matching or exact-answer evaluation is insufficient. To address this, we design a structured evaluation prompt that assesses responses along two complementary dimensions: \textit{relevance} and \textit{completeness}. 

The \textbf{relevance} score measures whether the response directly addresses the question and remains focused on the required information. This helps identify cases where a model produces unrelated or partially relevant descriptions. The \textbf{completeness} score evaluates how fully the response captures the key information present in the ground-truth answer. This metric is particularly important for long-video quality understanding tasks, where correct answers often require identifying multiple visual cues, distortions, or temporal events. By separating these two dimensions, the prompt allows us to distinguish between answers that are generally on-topic for long-video quality understanding but lack sufficient detail, and those that fully capture the necessary information. This design provides a more nuanced evaluation of open-ended responses compared to single-score correctness metrics. The corresponding relevance and completeness percentages are reported in Table~5 of the main paper.

\noindent
The full evaluation prompt used for scoring is shown below.
{\color{blue!70!black}
\begin{verbatim}
You are an evaluator for the Video Quality Understanding open-ended 
responses. Your task is to assess a respondent's answer against 
a question-answer pair in the dataset.

Evaluation Criteria
1. Completeness Score (0–1)
Evaluate how completely the response captures the information 
in the ground-truth answer.
0.0 → The response does not capture the key information from the
ground-truth answer.
0.5 → The response partially captures the information but misses 
important details.
1.0 → The response fully captures all essential information from 
the ground-truth answer.

2. Relevance Score (0–1)
Evaluate how relevant the response is to the question.
0.0 → Completely off-topic.
0.25 → Mostly irrelevant with only slight relation to the question.
0.5 → Partially relevant but contains unnecessary or 
unrelated content.
0.75 → Mostly relevant and focused on the question.
1.0 → Fully relevant and directly answers the question with 
no irrelevant content.

Input
Question: {question}
Answer (from dataset): {scoring_points}
Respondent Answer: {answer}

Output - return the final scores in JSON format:
{
  "completeness_score": <value between 0 and 1>,
  "relevance_score": <value between 0 and 1>
}
\end{verbatim}
}

\section{Challenges and Limitations}

While LongVQUBench provides a comprehensive benchmark for long-video quality understanding, several challenges and limitations remain. Handling long videos poses storage and computational challenges, making dataset processing and evaluation resource-intensive. Open-ended questions are difficult to evaluate reliably due to variability in phrasing and partial answers, resulting in lower completeness scores compared to MCQs. Additionally, the performance of video LVLMs varies across models and hierarchical levels, reflecting sensitivity to temporal reasoning and frame-level distortion detection. Finally, proprietary models often outperform open-source counterparts, limiting comprehensive comparison across all systems. These factors highlight areas for future improvement in model design, evaluation pipelines, and benchmark scalability.

\end{document}